%% file: main.tex
\documentclass[letterpaper]{article} %
\usepackage{aaai24}  %
\usepackage{times}  %
\usepackage{helvet}  %
\usepackage{courier}  %
\usepackage[hyphens]{url}  %
\usepackage{graphicx} %
\usepackage{natbib}  %
\usepackage{caption} %
\usepackage{algorithm}
\usepackage{algorithmic}

\usepackage{newfloat}
\usepackage{listings}
\usepackage{times}
\usepackage{epsfig}
\usepackage{amsmath}
\usepackage{amssymb}
\usepackage{booktabs}
\usepackage{multirow}
\usepackage{makecell}
\usepackage{rotating}
\usepackage{tabularray}
\usepackage{ragged2e}
\usepackage{array}
\usepackage{subcaption}
\usepackage{lipsum}
\usepackage[dvipsnames]{xcolor}
\usepackage{dsfont}
\usepackage{enumitem}
\usepackage{verbatim}
\usepackage{comment}
\usepackage{bm}
\usepackage{adjustbox}
\usepackage{pifont}
\usepackage{stmaryrd}
\usepackage{xfrac}
\usepackage{colortbl}
\usepackage{cuted}
\usepackage{subcaption}
\usepackage{hhline}
\usepackage{tabularx}
\usepackage[font=small, labelfont=bf]{caption}

\newcommand\gianni{\textcolor{black}}
\newcommand{\Xuanlong}{\textcolor{black}}

\DeclareMathOperator*{\argmin}{argmin}
\DeclareMathOperator*{\argmax}{argmax}

\definecolor{codeblue}{rgb}{0.3,0.55,0.78}
\definecolor{airforceblue}{rgb}{0.36, 0.54, 0.66}

\definecolor{lowerbetter}{RGB}{240,177,135}
\definecolor{higherbetter}{RGB}{199, 223, 181}

\definecolor{codegreen}{rgb}{0,0.6,0}
\definecolor{codegray}{rgb}{0.5,0.5,0.5}
\definecolor{codepurple}{rgb}{0.58,0,0.82}
\lstdefinestyle{mystyle}{
  commentstyle=\color{codegreen},
  keywordstyle=\color{magenta},
  numberstyle=\tiny\color{codegray},
  stringstyle=\color{codepurple},
  basicstyle=\ttfamily\footnotesize,
  breakatwhitespace=false,         
  breaklines=true,                 
  captionpos=b,                    
  keepspaces=true,                 
  numbers=left,                    
  numbersep=5pt,                  
  showspaces=false,                
  showstringspaces=false,
  showtabs=false,                  
  tabsize=2
}

\newcommand{\second}{\cellcolor[rgb]{0.847,0.882,0.949}}
\newcommand{\first}{\cellcolor[rgb]{0.557,0.663,0.859}}

\usepackage[breaklinks=true,bookmarks=false,citebordercolor={1 1 1},filebordercolor={1 1 1},linkbordercolor={1 1 1},citecolor=airforceblue,colorlinks=true]{hyperref}

\DeclareCaptionStyle{ruled}{labelfont=normalfont,labelsep=colon,strut=off} 
\floatstyle{ruled}
\newfloat{listing}{tb}{lst}{}
\floatname{listing}{Listing}
\definecolor{codegreen}{rgb}{0,0.6,0}
\definecolor{codegray}{rgb}{0.5,0.5,0.5}
\definecolor{codepurple}{rgb}{0.58,0,0.82}
\lstdefinestyle{mystyle}{
  commentstyle=\color{codegreen},
  keywordstyle=\color{magenta},
  numberstyle=\tiny\color{codegray},
  stringstyle=\color{codepurple},
  basicstyle=\ttfamily\footnotesize,
  breakatwhitespace=false,         
  breaklines=true,                 
  captionpos=b,                    
  keepspaces=true,                 
  numbers=left,                    
  numbersep=5pt,                  
  showspaces=false,                
  showstringspaces=false,
  showtabs=false,                  
  tabsize=2
}

\newcounter{relctr} 
\everydisplay\expandafter{\the\everydisplay\setcounter{relctr}{0}} 
\newcommand\labelrel[2]{%
  \begingroup
    \refstepcounter{relctr}%
    \stackrel{\textnormal{(\alph{relctr})}}{\mathstrut{#1}}%
    \originallabel{#2}%
  \endgroup
}
\AtBeginDocument{\let\originallabel\label} 

\title{Discretization-Induced Dirichlet Posterior for Robust Uncertainty Quantification on Regression}
\author{
    Xuanlong Yu\textsuperscript{\rm 1,\rm 2},
    Gianni Franchi\textsuperscript{\rm 2},
    Jindong Gu\textsuperscript{\rm 3},
    Emanuel Aldea\textsuperscript{\rm 1}
}
\affiliations{
\textsuperscript{\rm 1}SATIE, Paris-Saclay University\\
\textsuperscript{\rm 2}U2IS, ENSTA Paris, Institut Polytechnique de Paris\\
\textsuperscript{\rm 3}University of Oxford
}

\begin{document}

\maketitle

\begin{abstract}
Uncertainty quantification is critical for deploying deep neural networks (DNNs) in real-world applications.
An Auxiliary Uncertainty Estimator (AuxUE) is one of the most effective means to estimate the uncertainty of the main task prediction without modifying the main task model.
To be considered robust, an AuxUE must be capable of maintaining its performance and triggering higher uncertainties while encountering Out-of-Distribution (OOD) inputs, i.e., to provide robust aleatoric and epistemic uncertainty. However, for vision regression tasks, current AuxUE designs are mainly adopted for aleatoric uncertainty estimates, and AuxUE robustness has not been explored. 
In this work, we propose a generalized AuxUE scheme for more robust uncertainty quantification on regression tasks. 
Concretely, to achieve a more robust aleatoric uncertainty estimation, different distribution assumptions are considered for heteroscedastic noise, and Laplace distribution is finally chosen to approximate the prediction error.
For epistemic uncertainty, we propose a novel solution named Discretization-Induced Dirichlet pOsterior (DIDO), which models the Dirichlet posterior on the discretized prediction error.
Extensive experiments on age estimation, monocular depth estimation, and super-resolution tasks show that our proposed method can provide robust uncertainty estimates in the face of noisy inputs and that it can be scalable to both image-level and pixel-wise tasks. Code is available at \href{https://github.com/ENSTA-U2IS/DIDO}{https://github.com/ENSTA-U2IS/DIDO}.
\end{abstract}

\section{Introduction}
\label{sec:Introduction}

Uncertainty quantification in deep learning has gained significant attention in recent years~\cite{blundell2015weight,kendall2017uncertainties,lakshminarayanan2017simple,abdar2021review}. Deep Neural Networks (DNNs) frequently provide overconfident predictions and lack uncertainty estimates, especially for regression models outputting single point estimates, affecting the interpretability and credibility of the prediction results.

There are two types of uncertainty in DNNs: unavoidable aleatoric uncertainty caused by data noise, and reducible epistemic or knowledge uncertainty due to insufficient training data~\cite{hullermeier2021aleatoric,kendall2017uncertainties, malinin2018predictive}. Disentangling and estimating them can better guide the decision-making based on DNN predictions.
Many seminal methods~\cite{blundell2015weight, gal2016dropout, lakshminarayanan2017simple, kendall2017uncertainties, wen2020batchensemble, franchi2022latent} have been proposed to capture these two types of uncertainty. However, these methods require extensive modifications to the underlying model structure or more computational cost. Furthermore, since DNNs are often designed as task-oriented, obtaining uncertainty estimates by changing the structure of DNNs might reduce main task performance.

As one of the most effective methods, Auxiliary Uncertainty Estimators (AuxUE)~\cite{corbiere2019addressing, yu2021slurp, jain2021deup, corbiere2021beyond, besnier2021triggering, upadhyay2022bayescap, shen2022post} aim to obtain uncertainty estimates without affecting the main task performance. AuxUEs are DNNs that rely on the main task models used for estimating the uncertainty of the main task prediction. They are trained using the input, output, or intermediate features of the \textit{pre-trained} main task model.
In practice, the model inputs can be distribution-shifted from the training set, such as samples disturbed by noise~\cite{hendrycks2019benchmarking}, or even Out-of-Distribution (OOD) data.
{The pre-trained main task models mainly exhibit aleatoric uncertainty in the outputs given the In-Distribution (ID) inputs. Meanwhile, higher epistemic uncertainty is expected to be raised when OOD data is fed.}
A robust AuxUE is required in this case to provide robust aleatoric uncertainty estimates when facing In-Distribution (ID) inputs and epistemic uncertainty estimates when encountering OOD inputs. This can help to make effective decisions under anomalies and uncertainty~\cite{guo2022survey}, such as in autonomous driving~\cite{arnez2020comparison}.
Based on these requirements, 
the prerequisite for a robust AuxUE, thus, is to disentangle the two types of uncertainty. Disentangling can help estimate the epistemic uncertainty and find a more robust aleatoric uncertainty estimation solution.

For vision regression tasks, basic AuxUE addresses only aleatoric uncertainty estimation~\cite{yu2021slurp}. Recent works~\cite{upadhyay2022bayescap,qu2022improving} aim to improve the generalization ability of the basic AuxUEs. 
In DEUP~\cite{jain2021deup}, the authors propose to add a density estimator based on normalizing flows~\cite{rezende2015variational} in the AuxUE, yet challenging to apply on pixel-wise vision tasks.
In the current context, both the robustness analysis and modeling of epistemic uncertainty are underexplored for vision regression problems.

To further explore robust aleatoric and epistemic uncertainty estimation in vision regression tasks, in this work, we propose a novel uncertainty quantification solution based on AuxUE. For estimating aleatoric uncertainty, we follow the approach of previous works such as~\cite{nix1994uncertainty,kendall2017uncertainties,yu2021slurp,upadhyay2022bayescap} and model the heteroscedastic noise using different distribution assumptions. For epistemic uncertainty quantification, we apply a discretization approach to the continuous prediction errors of the main task. This helps to mitigate the numerical impact of the training targets, which may be distributed in a long-tailed manner. With the discretized prediction errors, we propose parameterizing Dirichlet posterior~\cite{sensoy2018evidential,charpentier2020posterior,joo2020being} for estimating epistemic uncertainty without relying on OOD data during the training process.

In summary, our contributions are as follows: (1) We propose a generalized AuxUE solution for aleatoric and epistemic uncertainty estimation; (2) We propose Discretization-Induced Dirichlet pOsterior (DIDO), a new epistemic uncertainty estimation strategy for regression{, which, to the best of our knowledge, is the only existing work employing this distribution for regression}; (3) We demonstrate that assuming the noise which affects the main task predictions to follow Laplace distribution can help AuxUE achieve a more robust aleatoric uncertainty estimation; (4) We propose a new evaluation strategy for the OOD analysis of pixel-wise regression tasks based on systematically non-annotated patterns.
We show the robustness and scalability of the proposed generalized AuxUE and DIDO on the age estimation, super-resolution and monocular depth estimation tasks.

\section{Related works}
\label{sec:related_works}
\input{1related_works}

\section{Method}
\label{sec:Method}
\input{2methods_v2.tex}

\section{Experiments}
\label{sec:Experiments}
\input{3_0experiments_newer}
\input{3_1experiments_newer}

\section{Conclusion}
In this paper, we propose a new solution for uncertainty quantification on regression problems based on a generalized AuxUE.
We design and implement the experiments based on four different regression problems. By modeling heteroscedastic noise using Laplace distribution, the proposed AuxUE can achieve more robust aleatoric uncertainty. Meanwhile, the novel DIDO solution in our AuxUE can provide better epistemic uncertainty estimation performance on both image-level and pixel-wise tasks. 

\clearpage

\section*{Acknowledgements}
We acknowledge the support of the Saclay-IA computing platform. We also thank M\u ad\u alina Olteanu for the thought-provoking discussion for the article.

\bibliography{references}  

\clearpage

\input{4supp}

\end{document}

%% file: 1related_works.tex
\paragraph{Auxiliary uncertainty estimation}
Auxiliary uncertainty estimation strategies can be divided into two categories: {unsupervised and supervised}. For the former, Dropout layer injection~\cite{mi2019training,gal2016dropout} samples the network by forward propagations, and \cite{hornauer2022gradient} proposed to use the gradients from the back-propagation. 
For the latter, AuxUEs are applied to obtain the uncertainty. {In addition to regression-oriented ones presented in Section~\ref{sec:Introduction}, we here introduce classification-oriented solutions.} ConfidNet~\cite{corbiere2019addressing} and KLoS~\cite{corbiere2021beyond} learn the true class probability and evidence for the DNNs, respectively. Shen et al.~\cite{shen2022post} apply evidential classification~\cite{joo2020being} to their AuxUE. ObsNet~\cite{besnier2021triggering} uses adversarial noise to provide more abundant training targets in semantic segmentation task for their AuxUE.

\paragraph{Evidential deep learning and Dirichlet networks}
Evidential deep learning~\cite{ulmer2021survey} (EDL) is a modern application of the Dempster-Shafer Theory~\cite{dempster1968generalization} to estimate epistemic uncertainty with single forward propagation. In classification tasks, EDL is usually formed as parameterizing a prior~\cite{malinin2018predictive,malinin2019reverse} or a posterior~\cite{joo2020being,charpentier2020posterior,natpn,sensoy2018evidential} Dirichlet distribution.
In regression problems, EDL estimates the parameters of the conjugate prior of Gaussian distribution~\cite{amini2020deep, natpn,malinin2020regression}.
Multi-task learning is recently applied to alleviate main task performance degradation due to applying such techniques~\cite{oh2022improving}, yet using AuxUE will not affect main task performance. Therefore, we apply EDL to our AuxUE. Moreover, we are the first to apply the Dirichlet network to the regression tasks by discretizing the main task prediction errors.

\paragraph{Robustness of uncertainty estimation}
A robust uncertainty estimator should show stable performance when encountering images perturbed to varying degrees~\cite{michaelis2019benchmarking,hendrycks2019benchmarking,kamann2021benchmarking}. Similar studies are applied to evaluate the robustness of uncertainty estimates~\cite{yeo2021robustness,franchi2022latent}. Meanwhile, it should provide a higher uncertainty when facing OOD data, such as in classification tasks~\cite{hendrycks17baseline,liang2018enhancing}. In image-level regression, we can use the definition of OOD from image classification~\cite{techapanurak2021practical} in, for example, age estimation task. But for pixel-wise regression tasks, the notion of OOD data is ill-defined. Typical OOD analysis estimates uncertainty on a different dataset than the training dataset~\cite{natpn}.
Yet, image patterns that are rarely assigned ground truth values in the training set can also be regarded as OOD. 
In this work, we also provide a new evaluation strategy for OOD patterns based on outdoor depth estimation to compensate for this experimental shortfall.

%% file: 2methods_v2.tex
\begin{figure*}[t]
\centering
\includegraphics[width=0.72\linewidth]{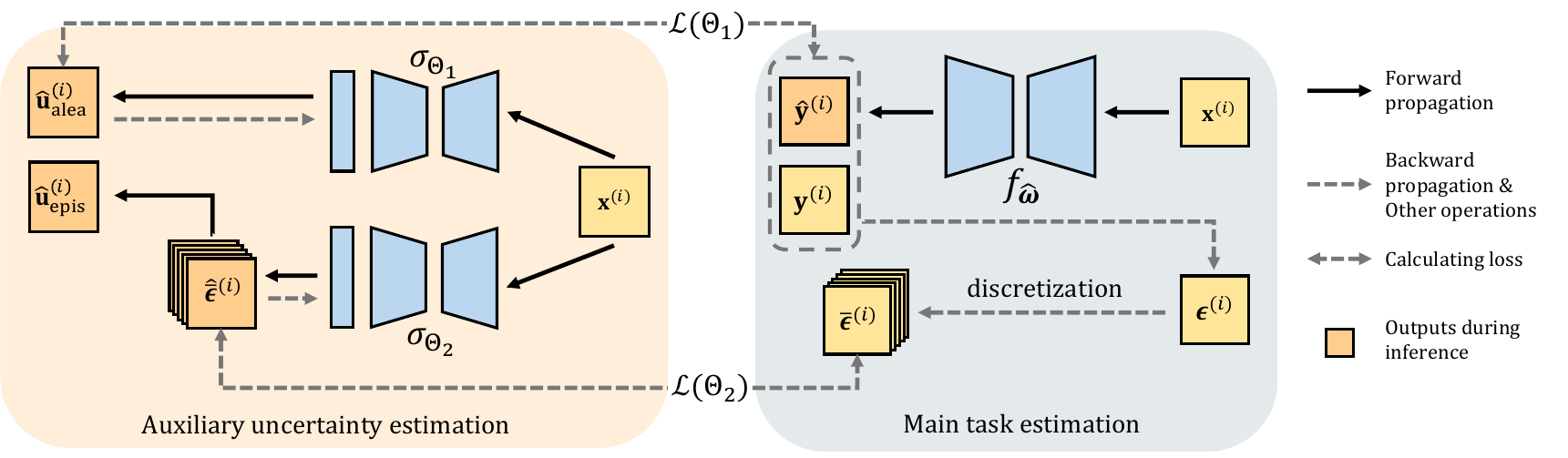}
\caption{\textbf{Pipeline of our proposed AuxUE solution.} A generalized AuxUE is considered with two DNNs $\sigma_{\Theta_1}$ and $\sigma_{\Theta_2}$ for estimating aleatoric and epistemic uncertainty, respectively. \Xuanlong{Presented notations are consistent with and described in Section~\ref{sec:Method}.} The encoder parts of both DNNs can be shared, we compare the performance in Section~\ref{sec:mde}. The input of AuxUE can be the input, output,
or intermediate features of $f_{\hat{\boldsymbol{\omega}}}$, we here simplify it to the image $\mathbf{x}^{(i)}$ for brevity.
}
\label{fig:procedure}
\end{figure*}

In this section, we will first provide the notations and the problem settings.
We define a training dataset 
$\mathcal{D} =\{\mathbf{x}^{(i)}, {y^{(i)}}\}_i^N$
where $N$ is the number of images. 
We consider that $\mathbf{x},\mathbf{y}$ are drawn from a joint distribution $P(\mathbf{x}, \mathbf{y})$.
A pipeline for the main task and auxiliary uncertainty estimation is shown in Fig.~\ref{fig:procedure}.
We define a main task DNN $f_{\boldsymbol{\omega}}$ with trainable parameters $\boldsymbol{\omega}$ as shown in the blue area in Fig.~\ref{fig:procedure}.
Similar to~\cite{blundell2015weight}, \Xuanlong{we view $f_{\boldsymbol{\omega}}$ as a probabilistic model $P({y}|\mathbf{x}, \boldsymbol{\omega})$ which follows a Gaussian distribution $\mathcal{N}({y}| \mu, {\sigma}^{2})$~\cite{bishop2006pattern}.}
The variable $\sigma^2$ represents the variance of the noise in the DNN's prediction, and the variable $\mu$ is the prediction $\hat{y} = f_{\boldsymbol{\omega}}(\mathbf{x})$ in this case. The noise is considered here to be homoscedastic as all data have the same noise.
The parameter $\boldsymbol{\omega}$ is optimized by maximizing the \gianni{log-}likelihood: 
$\hat{\boldsymbol{\omega}} = \argmax_{\boldsymbol{\omega}} \log \left(P(\mathcal{D} | \boldsymbol{\omega}) \right)$
which is often performed by minimizing Negative Log Likelihood (NLL) loss in practice.
With the above-mentioned Gaussian assumption on $\hat{{y}}$, the NLL loss optimizes with the same objective as the Mean Square Error loss~\cite{bishop2006pattern}, 
thus, only the prediction goal ${y}$ is considered, and the uncertainty modeling is absent in the main task model training objective. 

AuxUE aims to obtain this missing uncertainty estimation without modifying $\hat{\boldsymbol{\omega}}$. 
We consider two DNNs $\sigma_{{\Theta}_1}$ and $\sigma_{{\Theta}_2}$ in our generalized AuxUE with parameters $\Theta_1$ and $\Theta_2$, i.e., the two DNNs in the orange area of the Fig.~\ref{fig:procedure}.
$\sigma_{{\Theta}_1}$ is for estimating aleatoric uncertainty $\mathbf{u}_\text{alea}$, and $\sigma_{{\Theta}_2}$ is for estimating epistemic uncertainty $\mathbf{u}_\text{epis}$. The backbone of $\sigma_{{\Theta}_1}$ and $\sigma_{{\Theta}_2}$ are based on the basic AuxUEs such as ConfidNet~\cite{corbiere2019addressing}, BayesCap~\cite{upadhyay2022bayescap} and SLURP~\cite{yu2021slurp} depending on the tasks.
The input of AuxUE can be the input, output, or intermediate features of $f_{\hat{\boldsymbol{\omega}}}$ and it depends on the design of the basic AuxUEs, which is not the focus of this paper.
For brevity, we simplify the input of AuxUE to the image $\mathbf{x}$. We detail the inputs for different experiments in Supplementary material (Supp) Section~\ref{supp:A}.

\subsection{Aleatoric uncertainty estimation on AuxUE}\label{sec:method_alea}
Based on the preliminaries of the settings, we now start with the first AuxUE $\sigma_{\Theta_1}$, which addresses $\mathbf{u}_\text{alea}$ estimation problem as in SLURP and BayesCap.

\gianni{We consider the data-dependent noise~\cite{goldberg1997regression,bishop1996regression,nix1994uncertainty} follows $\mathcal{N}(0, \boldsymbol{\sigma}^2)$.}
Then we use the DNN $\sigma_{\Theta_1}$ to estimate the heteroscedastic aleatoric uncertainty ${\mathbf{u}}_{\text{alea}}$~\cite{nix1994uncertainty,kendall2017uncertainties}.
$\widehat{\Theta}_1$ and the loss function $L(\Theta_1)$ are given by:
\begin{small}
\begin{align}
&\widehat{\Theta}_1 \text{=} \argmax_{\Theta_1}P(\mathcal{D}|\hat{\boldsymbol{\omega}},\Theta_1) \text{=} \argmax_{\Theta_1} \sum^{N}_{i=1} \log(P({y}^{(i)}|\mathbf{x}^{(i)},\hat{\boldsymbol{\omega}},\Theta_1))\nonumber\\
&\mathcal{L}(\Theta_1) {=} \frac{1}{N}\sum_{i=1}^{N}\left[\frac{1}{2}\log( \sigma_{\Theta_1}({\mathbf{x}^{(i)}})) {+} \frac{({y}^{(i)} {-} f_{\hat{\boldsymbol{\omega}}}(\mathbf{x}^{(i)}))^2}{2\sigma_{\Theta_1}(\mathbf{x}^{(i)})}\right]
\label{eq:2}
\end{align}
\end{small}\noindent
The top of the $\sigma_{\Theta_1}$ is an exponential or Softplus function to maintain the output non-negative. The aleatoric uncertainty estimation will be: $\hat{u}^{(i)}_{\text{alea}} = \sigma_{\Theta_1}(\mathbf{x}^{(i)})$.
Minimizing $\mathcal{L}(\Theta_1)$ is also equivalent to making $\sigma_{\Theta_1}$ correctly predict the main task errors on the training set according to likelihood maximization. The errors set is denoted as $\boldsymbol{\epsilon} = \{ \epsilon^{(i)} \}_{i=1}^N = \{ (y^{(i)} - f_{\hat{\boldsymbol{\omega}}}(\mathbf{x}^{(i)}))^2 \}_{i=1}^N$.

Given the fact that distribution assumption on the noise affecting $\hat{{y}}$ can be different than Gaussian, e.g., Laplacian~\cite{marks1978detection} and Generalized Gaussian distribution~\cite{nadarajah2005generalized,upadhyay2022bayescap} also been considered in this work, the corresponding loss functions are provided in Supp Section~\ref{supp:B}. The objective remains unchanged: employing AuxUE to estimate and predict the component associated with aleatoric uncertainty using various distribution assumptions.
\gianni{
Perturbing input data in various ways with different types of noise makes it challenging to accurately identify the actual noise distribution. Relying on a single distribution assumption and loss function can affect the reliability of aleatoric uncertainty estimates. In Section~\ref{sec:mde}, we assess the impact of different distribution assumptions and losses on the robustness of these estimates.}

\subsection{Epistemic uncertainty estimation on AuxUE}\label{sec:didp}
Modeling AuxUEs as formalized in Eq.~\ref{eq:2} helps to estimate aleatoric uncertainty for $f_{\hat{\boldsymbol{\omega}}}$. Yet, taking this uncertainty prediction as an indicator for epistemic uncertainty is not methodologically grounded.
Evidential learning is considered to be an effective uncertainty estimation approach~\cite{ulmer2021survey}, which can capture epistemic uncertainty with a single pass as introduced in Section~\ref{sec:related_works}.
We thus take it as an alternative to implement on AuxUE. In regression tasks, DNN estimates the parameters of the conjugate prior of Gaussian distribution, such as Normal Inverse Gamma (NIG) distribution~\cite{amini2020deep}.
The training will make the model fall back onto a NIG prior for the rare samples by {attaching lower evidence to the samples with higher prediction errors using a regularization term in the loss function~\cite{amini2020deep}}.
Yet, long-tailed prediction errors make standard AuxUE more inclined to give high evidence for most data points, thereby reducing its ability to estimate epistemic uncertainty. Our experiments also confirmed this tendency.

In contrast to previous works, which consider the \textit{numerical value} of the prediction errors for both aleatoric and epistemic uncertainty estimation, we disentangle them and apply discretization to mitigate numerical bias from long-tailed prediction errors. Specifically, $\sigma_{\Theta_1}$ focuses on aleatoric uncertainty considering the \textit{numerical value} of prediction errors, while for epistemic uncertainty, $\sigma_{\Theta_2}$ will consider the \textit{value-free categories} of the prediction errors. 
Specifically, we propose Discretization-Induced Dirichlet pOsterior (DIDO), involves discretizing prediction errors and estimating a Dirichlet posterior based on the discrete errors. Further details are provided in the following sections.

\paragraph{3.2.1 Discretization on prediction errors}\label{sec:discritization}
To mitigate numerical bias due to imbalanced data in our prediction error estimation, we employ a balanced discretization approach. 
Discretization is widely applied in classification approaches for regression~\cite{yu_car}.
The popular discretization methods can be generally divided into handcrafted~\cite{cao2017estimating} and adaptive~\cite{bhat2021adabins}. The latter requires computationally expensive components like mini-ViT~\cite{dosovitskiy2020image} to extract global features. Thus, we discretize prediction errors in a handcrafted way.

For pixel-wise scenarios, discretization is applied using per-image prediction errors, and for other cases, such as image-level tasks and 1D signal estimation, we use per-dataset prediction errors. Details and \Xuanlong{demo-code can be found in Supp Section~\ref{supp:C1} and~\ref{supp:C2} respectively}.

We divide the set of errors $\boldsymbol{\epsilon}$, denoted in Section~\ref{sec:method_alea}, into $K$ subsets, where the $k$th subset is represented by the subscript $k$. 
To do this, we sort the errors in ascending order and create a new set, denoted by $\boldsymbol{\epsilon}^\prime$, with the same elements as $\boldsymbol{\epsilon}$. Then we divide $\boldsymbol{\epsilon}^\prime$ into $K$ subsets of equal size, represented by $\{ \boldsymbol{\epsilon}_k\}_{k=1}^K$. Each error value $\epsilon^{(i)}$ is then replaced by the index of its corresponding subset $k\in [1, K]$, and transformed into a one-hot vector, denoted by $\bar{\boldsymbol{\epsilon}}^{(i)}$, as the final training target. Specifically, the one-hot vector is defined as:
\begin{small}
\begin{align}
\bar{\boldsymbol{\epsilon}}^{(i)} = [\bar{\epsilon}_{1}^{(i)} \ldots \bar{\epsilon}_{k}^{(i)} \ldots \bar{\epsilon}_{K}^{(i)}]^{\text{T}} \in \mathds{R}^K\label{eq:4}
\end{align}
\end{small}\noindent
where $\bar{\epsilon}_{k}^{(i)}=1$ if $\epsilon^{(i)}$ belongs to the $k$th subset, and 0 otherwise. Each subset or bin represents a class of error severity.
This process creates a new dataset, denoted by $\Bar{\mathcal{D}} = \{\mathbf{x}^{(i)}, \Bar{\boldsymbol{\epsilon}}^{(i)}\}_i^N$, consisting of discretized prediction errors represented as one-hot vectors, which serves for training the epistemic uncertainty estimator $\sigma_{\Theta_2}$.

\paragraph{3.2.2 Modeling epistemic uncertainty using $\boldsymbol{\epsilon}$ in auxiliary uncertainty estimation}
In a Bayesian framework, given an input $\mathbf{x}$, the predictive uncertainty of a DNN is modeled by $P(y|\mathbf{x}, \mathcal{D})$. Since we have a trained main task DNN, and as proposed in~\cite{malinin2018predictive}, we assume a point-estimate of $\boldsymbol{\omega}$ (denoted as $\hat{\boldsymbol{\omega}}$), then we have:
\begin{small}
\begin{align}
P(\boldsymbol{\omega}|\mathcal{D}) = \delta(\boldsymbol{\omega} - \hat{\boldsymbol{\omega}}) \rightarrow
P(y|\mathbf{x}, \mathcal{D}) \approx P(y|\mathbf{x}, \hat{\boldsymbol{\omega}})
\label{eq:transform1_dido}
\end{align}
\end{small}\noindent
with $\delta$ being the Dirac function.

We follow the previous assumption, i.e., the prediction is drawn from a Gaussian distribution $\mathcal{N}(y| {\mu}, {\sigma}^2)$ and according to~\cite{amini2020deep}, we denote $\boldsymbol{\alpha}$ as the parameters of the prior distributions of $({\mu}, {\sigma}^{2})$ and we have $P(\mu, \sigma^2| \boldsymbol{\alpha}, \hat{\boldsymbol{\omega}}) = P(\mu|\sigma^2, \boldsymbol{\alpha}, \hat{\boldsymbol{\omega}})P(\sigma^2| \boldsymbol{\alpha},{\boldsymbol{\omega}}^*)$. After introducing $\boldsymbol{\alpha}$ and Eq.~\ref{eq:transform1_dido}, we can approximate $P(y|\mathbf{x}, \mathcal{D})$  as:
\begin{small}
\begin{align}
    P(y| \mathbf{x}, \mathcal{D}) &= \iint P(y | \mathbf{x}, \sigma^2) P(\sigma^2 | \boldsymbol{\omega}) P(\boldsymbol{\omega} | \mathcal{D}) d\sigma^2 d\boldsymbol{\omega} \nonumber\\
    &= \int P(y | \mathbf{x}, \sigma^2) P(\sigma^2 | \mathcal{D}) d\sigma^2 \nonumber\\
    &\approx \int P(y|\mathbf{x}, \sigma^{2}) P(\sigma^{2} | \mathbf{x}, \boldsymbol{\alpha}, \hat{\boldsymbol{\omega}}) d{\sigma^{2}}
    \label{eq:predictive_u_dido}
\end{align}
\end{small}\noindent
\Xuanlong{Detailed derivation can be found in Supp Section~\ref{supp:C3}.} 

\Xuanlong{We can consider $\epsilon$ to be drawn from a continuous distribution parameterized by $\sigma^2$. The discrepancy in variances $P(\sigma^2 | \mathcal{D})$ can describe epistemic uncertainty of the final prediction and the variational approach can be applied~\cite{joo2020being,malinin2018predictive}: $P( \sigma^2 | \mathbf{x}, \boldsymbol{\alpha}, \hat{\boldsymbol{\omega}}) \approx P(\sigma^2 | {\mathcal{D}})$. After discretization, we can transform the approximation to $P( \boldsymbol{\pi} | \mathbf{x}, \boldsymbol{\alpha}, \hat{\boldsymbol{\omega}}) \approx P(\boldsymbol{\pi} | \Bar{\mathcal{D}})$, with $\Bar{\mathcal{D}}$ defined as in Section 3.2.1, $\boldsymbol{\pi}$ the parameters of a discrete distribution and $\boldsymbol{\alpha}$ re-defined as the prior distribution parameters of this discrete distribution. In the next section, we omit $\hat{\boldsymbol{\omega}}$ and $\mathbf{x}$ for the sake of brevity. }

\paragraph{3.2.3 Dirichlet posterior for epistemic uncertainty}
According to the previous discussions on the epistemic uncertainty modeling and error discretization, we model Dirichlet posterior~\cite{sensoy2018evidential,joo2020being,natpn} on the discrete errors $\bar{\boldsymbol{\epsilon}}$ to achieve epistemic uncertainty on the main task.

Intuitively, we consider each one-hot prediction error $\bar{\boldsymbol{\epsilon}}^{(i)}$ to be drawn from a categorical distribution, and
$\boldsymbol{\pi}^{(i)} = (\pi^{(i)}_{1}, \ldots, \pi^{(i)}_{K})$ denotes the random variable over this distribution, where $\sum_{k=1}^{K}\pi^{(i)}_{k} = 1$ and $\pi^{(i)}_{k} \in [0,1] \text{ for } k\in\{1,...,K\}$.
The conjugate prior of categorical distribution is a Dirichlet distribution:
\begin{small}
\begin{equation}
P(\boldsymbol{\pi}^{(i)}|\boldsymbol{\alpha}^{(i)}) = \frac{\Gamma(S^{(i)})}{\prod_{k=1}^K \Gamma(\alpha^{(i)}_{k})}\prod_{k=1}^K {\pi^{(i)}_{k}}^{\alpha^{(i)}_{k} - 1}
\label{eq:5}
\end{equation}
\end{small}\noindent
with $\Gamma(\cdot)$ the Gamma function, $\boldsymbol{\alpha}^{(i)}$ positive concentration parameters of Dirichlet distribution and $S^{(i)} = \sum_{k=1}^{K}\alpha^{(i)}_{k}$ the Dirichlet strength.

To get access to the epistemic uncertainty, 
the categorical posterior $P(\boldsymbol{\pi}|\bar{\mathcal{D}})$ is needed, yet it is untractable. Approximating $P(\boldsymbol{\pi}|\bar{\mathcal{D}})$ using Monte-Carlo sampling~\cite{gal2016dropout} or ensembles~\cite{lakshminarayanan2017simple} 
comes with an increased computational cost. Instead, 
we adopt a variational way to learn a Dirichlet distribution in Eq.~\ref{eq:5} to approximate $P(\boldsymbol{\pi}|\bar{\mathcal{D}})$
as in~\cite{joo2020being}.
Here, $\sigma_{\Theta_2}$ outputs the concentration parameters $\boldsymbol{\alpha}$ of $P(\boldsymbol{\pi}|\boldsymbol{\alpha})$, and $\boldsymbol{\alpha}$ update according to the observed inputs. It can also be viewed as collecting the evidence $\boldsymbol{e}$ as a measure for supporting the classification decisions for each class~\cite{sensoy2018evidential}, akin to estimating the Dirichlet posterior.

Since the numbers of data points are identical for each class in $\bar{\mathcal{D}}$, {and no $\boldsymbol{e}^{(i)}$ output before training}, 
we set the initial $\boldsymbol{\alpha}$ as $\mathbf{1}$ so that the Dirichlet concentration parameters can be formed as in~\cite{sensoy2018evidential,charpentier2020posterior}: $\boldsymbol{\alpha}^{(i)} = \boldsymbol{e}^{(i)} + \mathbf{1} = \sigma_{\Theta_2}(\mathbf{x}^{(i)}) + \mathbf{1}$, where $\boldsymbol{e}^{(i)}$ is given by an exponential function on the top of $\sigma_{\Theta_2}$. Then we minimize the Kullback-Leibler (KL) divergence between the variational distribution $P(\boldsymbol{\pi}|\mathbf{x}, \Theta_2)$ and the true posterior $P(\boldsymbol{\pi}|\bar{\mathcal{D}})$ to achieve $\widehat{\Theta}_2$:
\begin{small}
\begin{align}
\widehat{\Theta}_2 = \argmin_{\Theta_2} &\text{ KL}[P(\boldsymbol{\pi}|\mathbf{x}, \Theta_2)||P(\boldsymbol{\pi}|\bar{\mathcal{D}})]\nonumber\\
= \argmin_{\Theta_2}&- \mathds{E}_{P(\boldsymbol{\pi}|\mathbf{x}, \Theta_2)}[\log P(\bar{\mathcal{D}}|\boldsymbol{\pi})] + \text{KL}[P(\boldsymbol{\pi}|\mathbf{x}, \Theta_2) || P(\boldsymbol{\pi})]\nonumber 
\end{align}
\end{small}\noindent
The loss function will be equivalent to minimizing the negative evidence lower bound~\cite{jordan1999introduction}, considering the prior distribution $P(\boldsymbol{\pi})$ as $\text{Dir}(\boldsymbol{1})$:
\begin{small}
\begin{align}
\mathcal{L}(\Theta_2) = \frac{1}{N}\sum^{N}_{i=1}\sum_{k=1}^{K}[\bar{\epsilon}^{(i)}_{k}(\psi(S^{(i)}) - \psi(\alpha_{k}^{(i)}))] \nonumber\\
+ \lambda \text{KL}(\text{Dir}(\boldsymbol{\alpha}^{(i)})||\text{Dir}(\boldsymbol{1}))\label{eq:loss_epis}
\end{align}
\end{small}\noindent
where $\psi$ is the digamma function, $\lambda$ is a positive hyperparameter for the regularization term and $\bar{\epsilon}$ is given by Eq.~\ref{eq:4}.

For measuring epistemic uncertainty, we consider using the spread in the
Dirichlet distribution~\cite{shen2022post,charpentier2020posterior}, which is shown in~\cite{shen2022post} to outperform other metrics, e.g. differential entropy. Specifically, the epistemic uncertainty is inversely proportional to the Dirichlet strength:
$\hat{u}^{(i)}_\text{epis} = \sigma_{\widehat{\Theta}_2}(\mathbf{x}^{(i)}) = \frac{K}{S^{(i)}}$.
The class corresponding to the maximum output from $\sigma_{\Theta_2}$ can also represent the aleatoric uncertainty. Yet, this is a rough estimate due to quantization errors and underperforming the other solutions. We provide the corresponding results in Supp Tab.~\ref{tab:alea_mde_supp}. 
Overall, we take only $\sigma_{\Theta_1}$ output as the aleatoric uncertainty. 
\\

In conclusion, we propose a generalized AuxUE with two components, namely $\sigma_{\Theta_1}$ and $\sigma_{\Theta_2}$, to quantify the uncertainty of the prediction given by the main task model. Based on different distribution assumptions on heteroscedastic noise in training data introduced in Section~\ref{sec:method_alea}, we can train $\sigma_{\Theta_1}$ to estimate aleatoric uncertainty. 
Meanwhile, as described in Section~\ref{sec:didp}, applying the proposed DIDO on $\sigma_{\Theta_2}$ and measuring the spread of Dirichlet distribution can help to estimate the epistemic uncertainty. 
Overall, we integrate the optimization for both uncertainty estimators, and the final loss for training the generalized AuxUE is:
\begin{small}
\begin{align}
\mathcal{L}_{\text{AuxUE}} = \mathcal{L}(\Theta_1) + \mathcal{L}(\Theta_2)
\end{align}
\end{small}\noindent
For $\mathcal{L}(\Theta_1)$, in addition to the Gaussian NLL, we will test other NLL loss functions according to different distribution assumptions in the experiment.

%% file: 3_0experiments_newer.tex
In this section, we first show the feasibility of the proposed generalized AuxUE on toy examples. Then, we demonstrate the effectiveness of epistemic uncertainty estimation using the proposed DIDO on age estimation and monocular depth estimation (MDE) tasks, and investigate the robustness of aleatoric uncertainty estimation on MDE task.  
Due to page limitations, the experiments for an example of OOD detection in tabular data regression and the super-resolution task are provided in Supp Section~\ref{supp:A2} and~\ref{supp:A4} respectively.

In the result tables, the top two performing methods are highlighted in color. All the results are averaged by three runs. The shar.enc$.$ and sep.enc$.$ denote respectively shared-parameters for the encoders and separate encoders of $\sigma_{\Theta_1}$ and $\sigma_{\Theta_2}$ in the generalized AuxUE. {For epistemic uncertainty, we compare our proposed method with the solutions based on modified main DNN: LDU~\cite{franchi2022latent}, Evidential learning (Evi.)~\cite{amini2020deep,joo2020being} and Deep Ensembles (DEns.)~\cite{lakshminarayanan2017simple}, as well as training-free methods: Gradient-based uncertainty (Grad.) ~\cite{hornauer2022gradient}, Variance based on Inject-Dropout (Inject.)~\cite{mi2019training}.}

The detailed implementations and the main task performance for all experiments are provided in Supp Section~\ref{supp:A}.

\begin{figure}[t]
        \begin{subfigure}{.234\textwidth}
            \includegraphics[width=\linewidth]{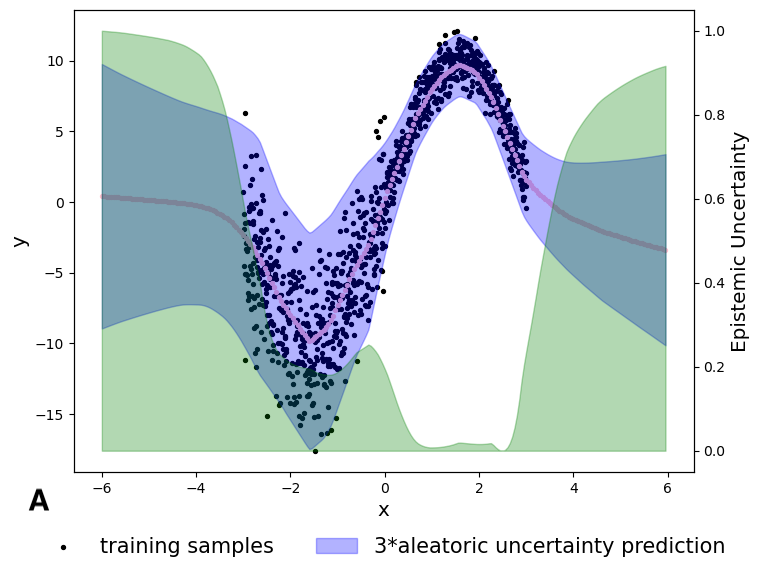}
        \end{subfigure}
        \hfill
        \begin{subfigure}{.234\textwidth}
            \includegraphics[width=\linewidth]{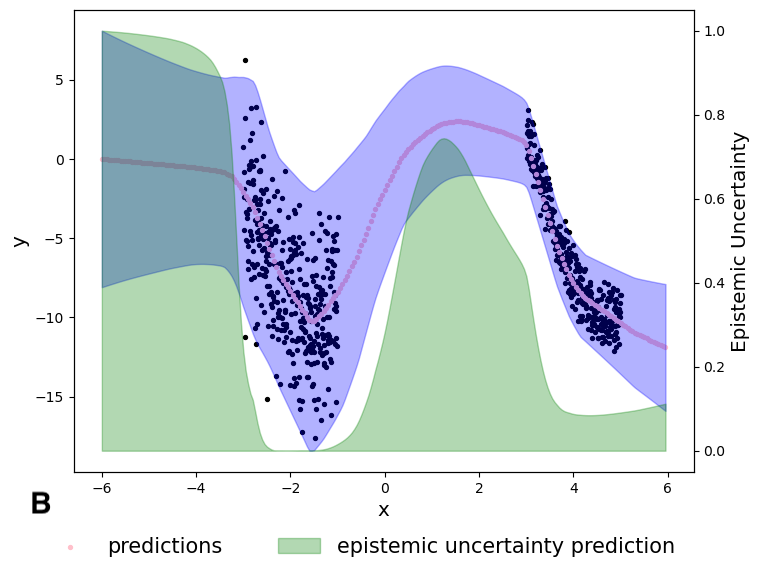}
        \end{subfigure}
\caption{\textbf{Results on 1D toy examples.} Aleatoric and epistemic uncertainty estimations given by our proposed AuxUE are presented respectively as the uncertainty interval and degree (0-1).}
\label{fig:toy}
\end{figure}

\subsection{Toy examples: Simple 1D regression}
We generate two toy datasets to illustrate uncertainty estimates given by our proposed AuxUE, as shown in Fig.~\ref{fig:toy}. 
In both examples, a tight aleatoric uncertainty estimation is provided on training data areas. For epistemic uncertainty, in Fig.~\ref{fig:toy}-A, DIDO provides small uncertainty until reaching the unknown inputs $x\notin[-3,3]$. In Fig.~\ref{fig:toy}-B, we report the `in-between' uncertainty estimates~\cite{foong2019between}. On the in-between part $x\in[-1,3]$, DIDO can provide higher epistemic uncertainty than in training set regions $x\in[-3,-1]$ and $x\in[3,5]$. In summary, the generalized AuxUE provides reliable uncertainty estimates in regions where training data is either present or absent.

\subsection{Age estimation and OOD detection}\label{sec:age_estimation}
Epistemic uncertainty estimation for age estimation is similar to one for classification problems but has rarely been discussed in previous works. {We use (unmodified) official ResNet34~\cite{he2016deep} checkpoints from Coral~\cite{cao2020rank} as the main task models. 
Our AuxUE is applied in a ConfidNet~\cite{corbiere2019addressing} style since it is more suitable for image-level tasks.}

\textbf{Evaluation settings and datasets\quad}We train the models on AFAD~\cite{Niu_2016_CVPR} training set and choose AFAD test set as the ID dataset for the OOD detection task. We 
take CIFAR10~\cite{cifar10}, SVHN~\cite{svhn}, MNIST~\cite{lecun1998mnist}, FashionMNIST~\cite{xiao2017fashion}, Oxford-Pets~\cite{parkhi12a} and Noise image generated by Pytorch~\cite{paszke2019pytorch} (FakeData) as the OOD datasets. We employ the Areas Under the receiver operating Characteristic (\textbf{\textit{AUC}}) and the Precision-Recall curve (\textbf{\textit{AUPR}}) (higher is better for both) to evaluate OOD detection performance.

\textbf{Results\quad}OOD detection results are shown in Tab.~\ref{tab:age}. DIDO performs the best on most datasets. The training-free methods also perform well, but we observe that the Gradient-based solution needs inversed uncertainty (inv.) to provide better performance. On the Pets dataset, DIDO performs worse than DEns. and aleatoric uncertainty estimation head $\sigma_{\Theta_1}$. We argue that images of pets provide features closer to facial information, resulting in higher evidence estimates given by DIDO. While $\sigma_{\Theta_1}$ performs better in this case, which can jointly make AuxUE a better uncertainty estimator. Overall, we consider that using generalized AuxUE with DIDO is an alternative that can better detect OOD inputs than ensembling-based solutions.

\begin{table}[t!]
        \centering
\scalebox{0.63}{
\begin{tabular}{llcclccclcc} 
\toprule
 &  & \multicolumn{2}{c}{AuxUE} & \multicolumn{1}{c}{} & \multicolumn{3}{c}{Modified main DNN} & \multicolumn{1}{c}{} & \multicolumn{2}{c}{Training-free} \\ 
\cmidrule[\heavyrulewidth]{3-4}\cmidrule[\heavyrulewidth]{6-8}\cmidrule[\heavyrulewidth]{10-11}
\begin{tabular}[c]{@{}l@{}}OOD\\Dataset\end{tabular} & Metrics & \begin{tabular}[c]{@{}c@{}}Ours~\\$\sigma_{\Theta_1}$\end{tabular} & \begin{tabular}[c]{@{}c@{}}Ours~$\sigma_{\Theta_2}$\\DIDO\end{tabular} &  & LDU & Evi. & DEns. &  & \begin{tabular}[c]{@{}c@{}}Grad.\\(inv.)\end{tabular} & Inject. \\ 
\toprule
\multirow{2}{*}{CIFAR10} & AUC~$\uparrow$ & 96.0 & {\cellcolor[rgb]{0.545,0.647,0.839}}\textbf{100} &  & 95.2 & 50.0 & {\cellcolor[rgb]{0.847,0.882,0.949}}99.2 &  & {\cellcolor[rgb]{0.545,0.647,0.839}}\textbf{100} & 94.5 \\
 & AUPR~$\uparrow$ & 91.7 & {\cellcolor[rgb]{0.545,0.647,0.839}}\textbf{100} &  & 88.3 & 23.4 & {\cellcolor[rgb]{0.847,0.882,0.949}}95.1 &  & {\cellcolor[rgb]{0.545,0.647,0.839}}\textbf{100} & 87.3 \\ 
\midrule
\multirow{2}{*}{SVHN} & AUC $\uparrow$ & 98.3 & {\cellcolor[rgb]{0.545,0.647,0.839}}\textbf{100} &  & 94.8 & 50.0 & {\cellcolor[rgb]{0.847,0.882,0.949}}99.2 &  & {\cellcolor[rgb]{0.545,0.647,0.839}}\textbf{100} & 94.0 \\
 & AUPR $\uparrow$ & {\cellcolor[rgb]{0.847,0.882,0.949}}98.1 & {\cellcolor[rgb]{0.545,0.647,0.839}}\textbf{100} &  & 93.2 & 44.3 & 97.8 &  & {\cellcolor[rgb]{0.545,0.647,0.839}}\textbf{100} & 92.5 \\ 
\midrule
\multirow{2}{*}{MNIST} & AUC $\uparrow$ & 97.8 & {\cellcolor[rgb]{0.545,0.647,0.839}}\textbf{100} &  & 97.6 & 50.0 & {\cellcolor[rgb]{0.847,0.882,0.949}}99.6 &  & {\cellcolor[rgb]{0.545,0.647,0.839}}\textbf{100} & 98.8 \\
 & AUPR $\uparrow$ & 93.9 & {\cellcolor[rgb]{0.545,0.647,0.839}}\textbf{100} &  & 93.8 & 23.4 & {\cellcolor[rgb]{0.847,0.882,0.949}}97.2 &  & {\cellcolor[rgb]{0.545,0.647,0.839}}\textbf{100} & 96.9 \\ 
\midrule
\multirow{2}{*}{\begin{tabular}[c]{@{}l@{}}Fashion \\MNIST~\end{tabular}} & AUC $\uparrow$ & 97.7 & {\cellcolor[rgb]{0.545,0.647,0.839}}\textbf{100} &  & 95.6 & 50.0 & {\cellcolor[rgb]{0.847,0.882,0.949}}99.1 &  & {\cellcolor[rgb]{0.545,0.647,0.839}}\textbf{100} & 97.7 \\
 & AUPR $\uparrow$ & 94.0 & {\cellcolor[rgb]{0.545,0.647,0.839}}\textbf{100} &  & 89.3 & 23.4 & 93.8 &  & {\cellcolor[rgb]{0.545,0.647,0.839}}\textbf{100} & {\cellcolor[rgb]{0.847,0.882,0.949}}94.2 \\ 
\midrule
\multirow{2}{*}{\begin{tabular}[c]{@{}l@{}}Oxford\\Pets\end{tabular}} & AUC $\uparrow$ & {\cellcolor[rgb]{0.545,0.647,0.839}}\textbf{82.9} & 55.9 &  & 31.5 & 50.1 & {\cellcolor[rgb]{0.847,0.882,0.949}}56.1 &  & 50.7 & 48.6 \\
 & AUPR $\uparrow$ & {\cellcolor[rgb]{0.545,0.647,0.839}}\textbf{53.3} & {\cellcolor[rgb]{0.847,0.882,0.949}}23.9 &  & 12.5 & 18.5 & 21.3 &  & 19.6 & 20.3 \\ 
\midrule
\multirow{2}{*}{\begin{tabular}[c]{@{}l@{}}Fake\\Data\end{tabular}} & AUC $\uparrow$ & 67.0 & {\cellcolor[rgb]{0.545,0.647,0.839}}\textbf{80.8} &  & {\cellcolor[rgb]{0.847,0.882,0.949}}70.0 & 50.0 & 33.2 &  & 45.9 & 45.1 \\
 & AUPR $\uparrow$ & {\cellcolor[rgb]{0.847,0.882,0.949}}59.7 & {\cellcolor[rgb]{0.545,0.647,0.839}}\textbf{70.2} &  & 58.8 & 49.5 & 37.8 &  & 46.3 & 44.6 \\
\bottomrule
\end{tabular}
}
        \caption{\textbf{OOD detection results on Age estimation task.} ID data is from Asian Face Age Dataset (AFAD)~\cite{Niu_2016_CVPR}.}
        \label{tab:age}
\end{table}

\begin{table}[t]
\centering
\arrayrulecolor{black}
\scalebox{0.61}{
\begin{tabular}{clcccccc}
\toprule
S & \multicolumn{1}{c}{Metrics} & Original & + Ggau & + Sgau & + NIG & \begin{tabular}[c]{@{}c@{}}Ours~(+ Lap)\\shar. enc. $\sigma_{\Theta_1}$\end{tabular} & \begin{tabular}[c]{@{}c@{}}Ours~(+ Lap)\\sep. enc. $\sigma_{\Theta_1}$\end{tabular} \\ 
\hline
\multirow{4}{*}{0} & AUSE-REL $\downarrow$ & {\cellcolor[rgb]{0.851,0.882,0.949}}0.013 & 0.014 & {\cellcolor[rgb]{0.851,0.882,0.949}}0.013 & {\cellcolor[rgb]{0.557,0.663,0.859}}\textbf{0.012} & {\cellcolor[rgb]{0.851,0.882,0.949}}0.013 & {\cellcolor[rgb]{0.851,0.882,0.949}}0.013 \\
 & AUSE-RMSE $\downarrow$ & 0.204 & 0.258 & {\cellcolor[rgb]{0.557,0.663,0.859}}\textbf{0.202} & 0.208 & 0.205 & {\cellcolor[rgb]{0.851,0.882,0.949}}0.203 \\
 & AURG-REL $\uparrow$ & {\cellcolor[rgb]{0.851,0.882,0.949}}0.023 & {\cellcolor[rgb]{0.851,0.882,0.949}}0.023 & {\cellcolor[rgb]{0.851,0.882,0.949}}0.023 & {\cellcolor[rgb]{0.557,0.663,0.859}}\textbf{0.024} & {\cellcolor[rgb]{0.851,0.882,0.949}}0.023 & {\cellcolor[rgb]{0.851,0.882,0.949}}0.023 \\
 & AURG-RMSE $\uparrow$ & 1.869 & 1.815 & {\cellcolor[rgb]{0.557,0.663,0.859}}\textbf{1.871} & 1.865 & 1.869 & {\cellcolor[rgb]{0.851,0.882,0.949}}1.870 \\ 
\midrule
\multirow{4}{*}{1} & AUSE-REL $\downarrow$ & {\cellcolor[rgb]{0.851,0.882,0.949}}0.019 & 0.021 & {\cellcolor[rgb]{0.851,0.882,0.949}}0.019 & {\cellcolor[rgb]{0.557,0.663,0.859}}\textbf{0.018} & {\cellcolor[rgb]{0.557,0.663,0.859}}\textbf{0.018} & {\cellcolor[rgb]{0.851,0.882,0.949}}0.019 \\
 & AUSE-RMSE $\downarrow$ & 0.340 & 0.482 & {\cellcolor[rgb]{0.557,0.663,0.859}}\textbf{0.332} & {\cellcolor[rgb]{0.851,0.882,0.949}}0.335 & {\cellcolor[rgb]{0.557,0.663,0.859}}\textbf{0.332} & 0.336 \\
 & AURG-REL $\uparrow$ & {\cellcolor[rgb]{0.851,0.882,0.949}}0.031 & 0.029 & {\cellcolor[rgb]{0.851,0.882,0.949}}0.031 & {\cellcolor[rgb]{0.557,0.663,0.859}}\textbf{0.032} & {\cellcolor[rgb]{0.557,0.663,0.859}}\textbf{0.032} & {\cellcolor[rgb]{0.851,0.882,0.949}}0.031 \\
 & AURG-RMSE $\uparrow$ & 2.357 & 2.215 & {\cellcolor[rgb]{0.557,0.663,0.859}}\textbf{2.365} & {\cellcolor[rgb]{0.851,0.882,0.949}}2.362 & {\cellcolor[rgb]{0.557,0.663,0.859}}\textbf{2.365} & 2.361 \\ 
\midrule
\multirow{4}{*}{2} & AUSE-REL $\downarrow$ & 0.024 & 0.026 & {\cellcolor[rgb]{0.851,0.882,0.949}}0.023 & {\cellcolor[rgb]{0.557,0.663,0.859}}\textbf{0.022} & {\cellcolor[rgb]{0.557,0.663,0.859}}\textbf{0.022} & {\cellcolor[rgb]{0.851,0.882,0.949}}0.023 \\
 & AUSE-RMSE $\downarrow$ & 0.483 & 0.707 & {\cellcolor[rgb]{0.557,0.663,0.859}}\textbf{0.463} & 0.479 & {\cellcolor[rgb]{0.851,0.882,0.949}}0.464 & 0.468 \\
 & AURG-REL $\uparrow$ & {\cellcolor[rgb]{0.851,0.882,0.949}}0.038 & 0.035 & {\cellcolor[rgb]{0.557,0.663,0.859}}\textbf{0.039} & {\cellcolor[rgb]{0.557,0.663,0.859}}\textbf{0.039} & {\cellcolor[rgb]{0.557,0.663,0.859}}\textbf{0.039} & {\cellcolor[rgb]{0.851,0.882,0.949}}0.038 \\
 & AURG-RMSE $\uparrow$ & 2.759 & 2.535 & {\cellcolor[rgb]{0.557,0.663,0.859}}\textbf{2.779} & 2.763 & {\cellcolor[rgb]{0.851,0.882,0.949}}2.777 & 2.774 \\ 
\midrule
\multirow{4}{*}{3} & AUSE-REL $\downarrow$ & {\cellcolor[rgb]{0.851,0.882,0.949}}0.033 & 0.036 & {\cellcolor[rgb]{0.557,0.663,0.859}}\textbf{0.031} & {\cellcolor[rgb]{0.557,0.663,0.859}}\textbf{0.031} & {\cellcolor[rgb]{0.557,0.663,0.859}}\textbf{0.031} & {\cellcolor[rgb]{0.557,0.663,0.859}}\textbf{\textbf{0.031}} \\
 & AUSE-RMSE $\downarrow$ & 0.795 & 1.176 & {\cellcolor[rgb]{0.851,0.882,0.949}}0.737 & 0.806 & 0.749 & {\cellcolor[rgb]{0.557,0.663,0.859}}\textbf{\textbf{0.730}} \\
 & AURG-REL $\uparrow$ & {\cellcolor[rgb]{0.851,0.882,0.949}}0.047 & 0.044 & {\cellcolor[rgb]{0.557,0.663,0.859}}\textbf{0.049} & {\cellcolor[rgb]{0.557,0.663,0.859}}\textbf{0.049} & {\cellcolor[rgb]{0.557,0.663,0.859}}\textbf{0.049} & {\cellcolor[rgb]{0.557,0.663,0.859}}\textbf{\textbf{0.049}} \\
 & AURG-RMSE $\uparrow$ & 3.243 & 2.862 & {\cellcolor[rgb]{0.851,0.882,0.949}}3.301 & 3.232 & 3.289 & {\cellcolor[rgb]{0.557,0.663,0.859}}\textbf{\textbf{3.308}} \\ 
\midrule
\multirow{4}{*}{4} & AUSE-REL $\downarrow$ & 0.056 & 0.057 & {\cellcolor[rgb]{0.851,0.882,0.949}}0.050 & 0.053 & 0.051 & {\cellcolor[rgb]{0.557,0.663,0.859}}\textbf{\textbf{0.049}} \\
 & AUSE-RMSE $\downarrow$ & 1.517 & 2.380 & {\cellcolor[rgb]{0.851,0.882,0.949}}1.364 & 1.582 & 1.430 & {\cellcolor[rgb]{0.557,0.663,0.859}}\textbf{\textbf{1.268}} \\
 & AURG-REL $\uparrow$ & 0.051 & 0.051 & {\cellcolor[rgb]{0.851,0.882,0.949}}0.058 & 0.054 & 0.056 & {\cellcolor[rgb]{0.557,0.663,0.859}}\textbf{\textbf{0.059}} \\
 & AURG-RMSE $\uparrow$ & 3.680 & 2.817 & {\cellcolor[rgb]{0.851,0.882,0.949}}3.834 & 3.615 & 3.767 & {\cellcolor[rgb]{0.557,0.663,0.859}}\textbf{\textbf{3.929}} \\ 
\midrule
\multirow{4}{*}{5} & AUSE-REL $\downarrow$ & 0.071 & 0.082 & {\cellcolor[rgb]{0.851,0.882,0.949}}0.064 & 0.069 & 0.066 & {\cellcolor[rgb]{0.557,0.663,0.859}}\textbf{\textbf{0.059}} \\
 & AUSE-RMSE $\downarrow$ & 2.202 & 3.878 & {\cellcolor[rgb]{0.851,0.882,0.949}}2.043 & 2.414 & 2.157 & {\cellcolor[rgb]{0.557,0.663,0.859}}\textbf{\textbf{1.760}} \\
 & AURG-REL $\uparrow$ & 0.056 & 0.045 & {\cellcolor[rgb]{0.851,0.882,0.949}}0.063 & 0.057 & 0.061 & {\cellcolor[rgb]{0.557,0.663,0.859}}\textbf{0.067} \\
 & AURG-RMSE $\uparrow$ & 4.054 & 2.377 & {\cellcolor[rgb]{0.851,0.882,0.949}}4.213 & 3.842 & 4.098 & {\cellcolor[rgb]{0.557,0.663,0.859}}\textbf{4.496} \\
\bottomrule
\end{tabular}
}
\caption{\textbf{Aleatoric uncertainty estimation results on Monocular Depth Estimation.}  $S=0$ represents original KITTI dataset and $S>0$ represents KITTI-C datasets.}
\label{tab:alea}
\end{table}

%% file: 3_1experiments_newer.tex
\subsection{Monocular depth estimation task}\label{sec:mde}

\begin{figure}[t]
\centering
        \begin{subfigure}{.234\textwidth}
            \includegraphics[width=\linewidth]{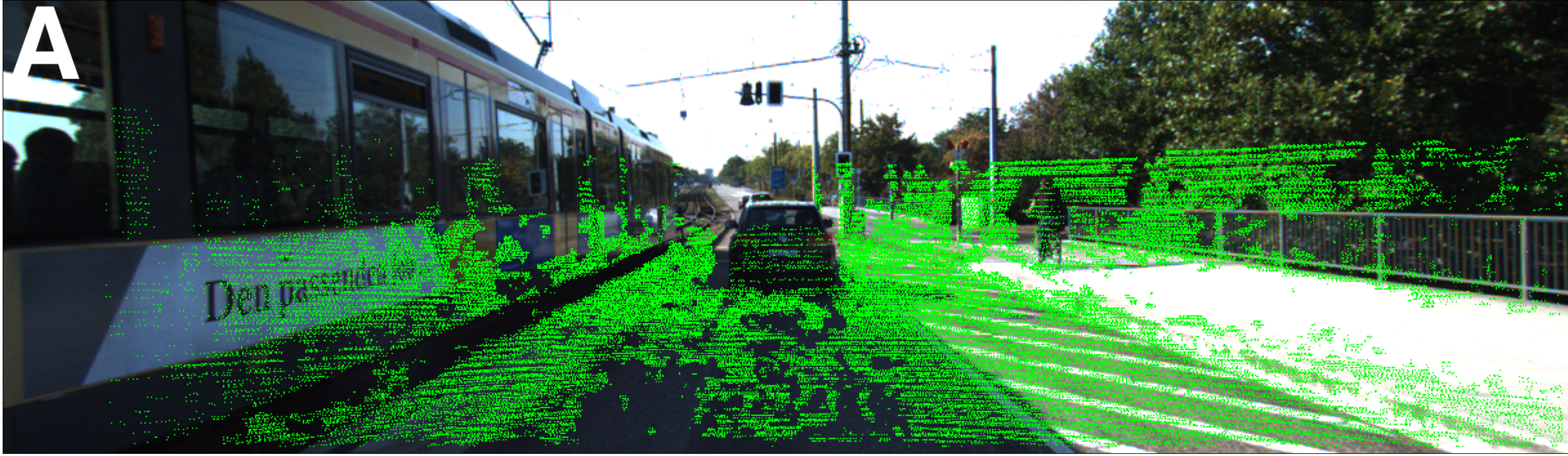}
        \end{subfigure}
        \begin{subfigure}{.234\textwidth}
            \includegraphics[width=\linewidth]{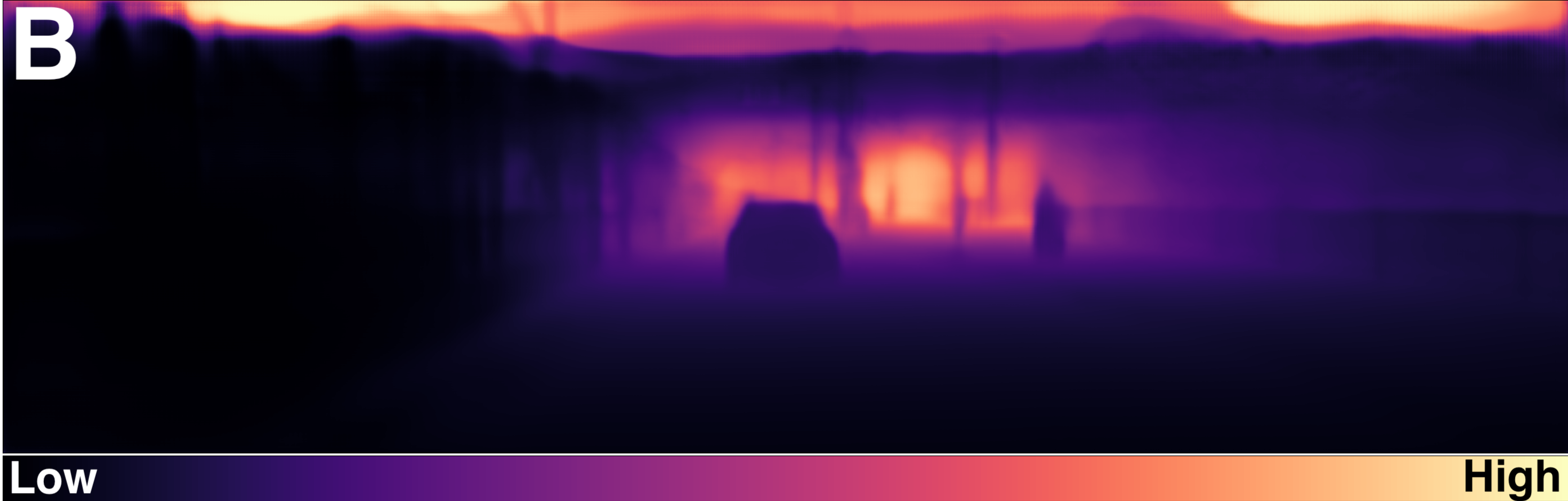}
        \end{subfigure}
        \begin{subfigure}{.234\textwidth}
            \includegraphics[width=\linewidth]{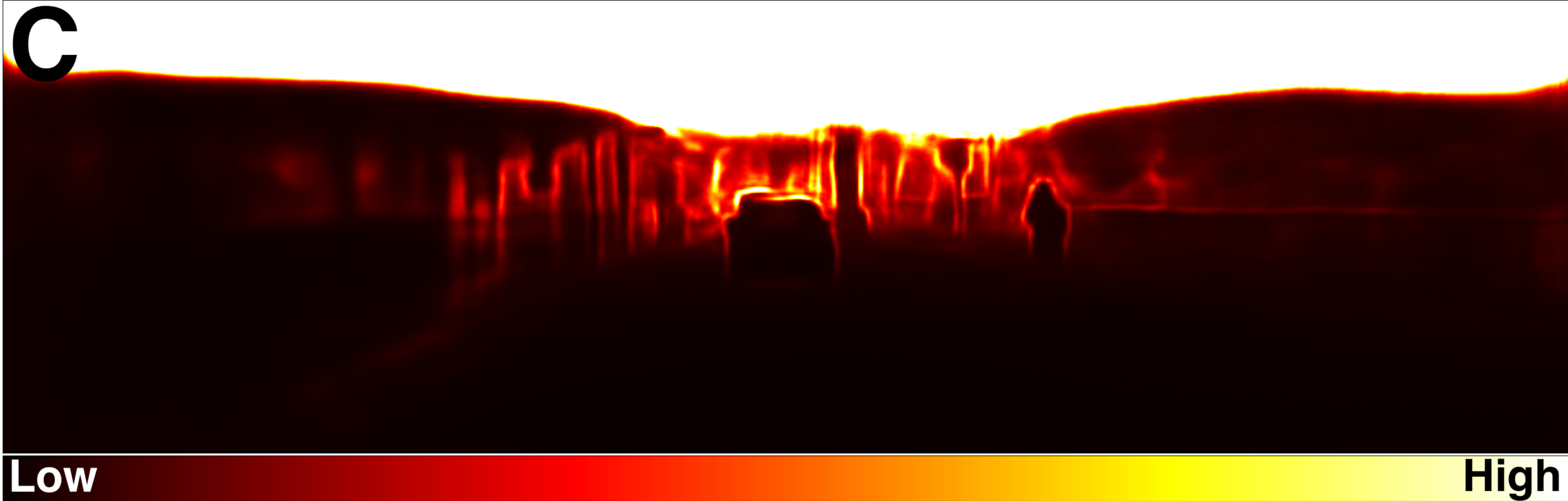}
        \end{subfigure}
        \begin{subfigure}{.234\textwidth}
            \includegraphics[width=\linewidth]{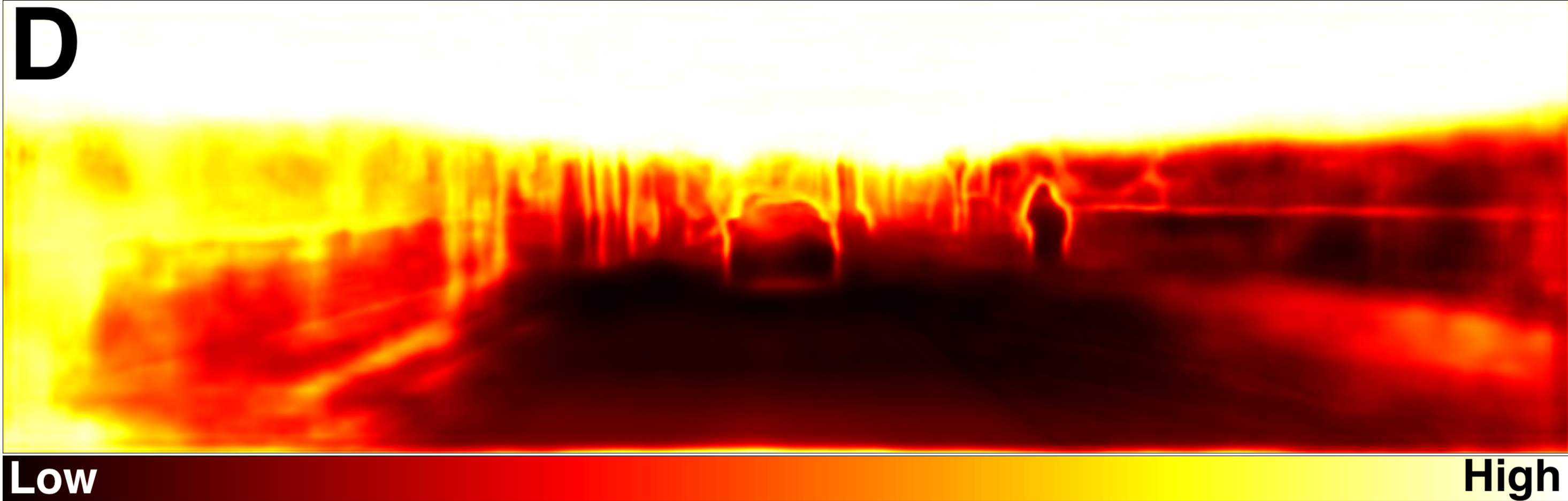}
        \end{subfigure}
\caption{\textbf{Illustrations of uncertainty estimations for MDE task.} A: input image, green points represent pixels with depth groundtruth; B: depth prediction; C and D: aleatoric and epistemic uncertainty estimations. The areas lacking depth groundtruth, e.g. sky and tramway, are assigned high uncertainty using DIDO.}
\label{fig:vis_mde}
\end{figure}

For the MDE task, we will evaluate both aleatoric and epistemic uncertainty estimation performance based on the AuxUE SLURP~\cite{yu2021slurp}. Our generalized AuxUE is also constructed using SLURP as the backbone. We use BTS~\cite{lee2019big} as the main task model and KITTI~\cite{Geiger2013IJRR, Uhrig2017THREEDV} Eigen-split~\cite{eigen2014depth} training set for training both BTS and AuxUE models. 

\subsubsection{4.3.1 Aleatoric uncertainty estimation}
In this section, the goal is to analyze the fundamental performance and robustness of aleatoric uncertainty estimation under different distribution assumptions. We choose simple Gaussian (Sgau)~\cite{nix1994uncertainty}, Laplacian (Lap), Generalized Gaussian (Ggau)~\cite{upadhyay2022bayescap} and Normal-Inverse-Gamma (NIG)~\cite{amini2020deep} distributions. We modify the loss functions and the head of the SLURP to output the desired parameters of the distributions. 

\textbf{Evaluation settings and datasets\quad}We first build Sparsification curves (SC)~\cite{bruhn2006confidence}: we achieve predictive SC by computing the prediction error of the remaining pixels after removing a certain partition of pixels (5$\%$ in our experiment) each time according to the highest uncertainty estimations.
We can also obtain an Oracle SC by removing the pixels according to the highest prediction errors. Then, we have the same metrics used in~\cite{poggi2020uncertainty}:  
Area Under the Sparsification Error (\textbf{\textit{AUSE}}, lower is better), and Area Under the Random Gain (\textbf{\textit{AURG}}, higher is better). We choose absolute relative error (REL) and root mean square error (RMSE) as the prediction error metrics.

We generate KITTI-C from KITTI Eigen-split validation set using the code of ImageNet-C~\cite{hendrycks2019benchmarking} to have different corruptions on the images to check the robustness of the uncertainty estimation solutions. We apply eighteen perturbations with five severities, including Gaussian noise, shot noise, etc., and take it along with the original KITTI for evaluation.

\textbf{Results\quad}
As shown in Tab.~\ref{tab:alea}, the Laplace assumption is more robust when the severity increases, while Gaussian one works better when the noise severity is smaller. We also check the proposed generalized AuxUE with a shared encoder. It shows that the epistemic uncertainty estimation branch affects the robustness of aleatoric uncertainty estimation in this case, especially under stronger noise.\\

The next sections show epistemic uncertainty estimation results based on different methods. Furthermore, in Supp Tab.~\ref{tab:epismde1_supp} and Tab.~\ref{tab:epismde2_supp}, we also verify whether aleatoric uncertainty methods based on different distribution assumptions
can generalize to the OOD data, i.e., provide high uncertainty to the unseen patterns, even without explicitly modeling epistemic uncertainty.

\paragraph{4.3.2 Robustness under dataset change}
This experiment will explore the predictive uncertainty performance encountering the dataset change. Supervised MDE is an ill-posed problem that heavily depends on the training dataset. In our case, the main task model is trained on the KITTI dataset, so the model will output meaningless results on the indoor data, which should trigger a high uncertainty estimation. The results are shown in Tab.~\ref{tab:epismde1}.

\textbf{Evaluation settings and datasets\quad}We take \textbf{\textit{AUC}} and \textbf{\textit{AUPR}} as evaluation metrics. We take all the valid pixels from the KITTI validation set  (ID) as the negative samples and the valid pixels from the NYU~\cite{Silberman:ECCV12} validation set (OOD) as the positive samples.

\textbf{Results\quad}Tab.~\ref{tab:epismde1} shows whether different uncertainty estimators can give correct indications facing the dataset change. Generalized Gaussian and Gradient-based methods can provide competitive results, while our method, especially DIDO, provides the best performance. 

\begin{table}[t]
\centering
\scalebox{0.69}{
\begin{tabular}{lccccccccc} \toprule
 & \multicolumn{2}{c}{\small{AuxUE with DIDO}} &  & \multicolumn{3}{c}{\small{Modified main DNN}} &  & \multicolumn{2}{c}{\small{Training-free}} \\ \cmidrule[\heavyrulewidth]{2-3}\cmidrule[\heavyrulewidth]{5-7}\cmidrule[\heavyrulewidth]{9-10}
Metrics & \begin{tabular}[c]{@{}c@{}}Ours $\sigma_{\Theta_2}$\\sep. enc.\end{tabular} & \begin{tabular}[c]{@{}c@{}}Ours $\sigma_{\Theta_2}$\\shar. enc.\end{tabular} &  & LDU & Evi. & DEns. &  & Grad. & Inject. \\ \toprule
{AUC}~$\uparrow$ & {\cellcolor[rgb]{0.843,0.882,0.949}}{98.1} & {\cellcolor[rgb]{0.557,0.667,0.863}}\textbf{98.4} &  & 58.1 & 70.6 & 62.1 &  & 78.4 & 18.3 \\
{AUPR}~$\uparrow$ & {\cellcolor[rgb]{0.843,0.882,0.949}}{99.3} & {\cellcolor[rgb]{0.557,0.667,0.863}}\textbf{99.4} &  & 79.5 & 77.8 & 76.7 &  & 92.6 & 62.3 \\ \bottomrule
\end{tabular}
}
\caption{\textbf{Epistemic uncertainty estimation results encountering dataset change on Monocular depth estimation task.} The evaluation dataset used here is NYU indoor depth dataset. }
\label{tab:epismde1}
\end{table}

\begin{table}[t]
\centering
\scalebox{0.65}{
\begin{tabular}{clccccccccc} 
\toprule
 &  & \multicolumn{2}{c}{AuxUE with DIDO} &  & \multicolumn{3}{c}{Modified main DNN} &  & \multicolumn{2}{c}{Training-free} \\ 
\cmidrule[\heavyrulewidth]{3-4}\cmidrule[\heavyrulewidth]{6-8}\cmidrule[\heavyrulewidth]{10-11}
S & Metrics & \begin{tabular}[c]{@{}c@{}}Ours $\sigma_{\Theta_2}$\\sep. enc.\end{tabular} & \begin{tabular}[c]{@{}c@{}}Ours $\sigma_{\Theta_2}$\\shar. enc.\end{tabular} &  & LDU & Evi. & DEns. &  & \begin{tabular}[c]{@{}c@{}}Grad.\\(inv.)\end{tabular} & Inject. \\ 
\toprule
\multirow{3}{*}{0} & AUC~$\uparrow$ & {\cellcolor[rgb]{0.545,0.647,0.839}}\textbf{100.0} & {\cellcolor[rgb]{0.847,0.882,0.949}}99.9 &  & 96.5 & 76.7 & 93.5 &  & 85.6 & 58.4 \\
 & AUPR~$\uparrow$ & {\cellcolor[rgb]{0.545,0.647,0.839}}\textbf{100.0} & {\cellcolor[rgb]{0.847,0.882,0.949}}99.0 &  & 93.8 & 42.6 & 70.0 &  & 76.3 & 28.1 \\
 & Sky-All~$\downarrow$ & 0.015 & 0.018 &  & 0.278 & 0.986 & {\cellcolor[rgb]{0.847,0.882,0.949}}0.005 &  & {\cellcolor[rgb]{0.545,0.647,0.839}}\textbf{0.001} & 0.800 \\ 
\midrule
\multirow{3}{*}{1} & AUC~$\uparrow$ & {\cellcolor[rgb]{0.545,0.647,0.839}}\textbf{100.0} & {\cellcolor[rgb]{0.847,0.882,0.949}}99.9 &  & 96.3 & 69.7 & 92.8 &  & 76.9 & 58.5 \\
 & AUPR~$\uparrow$ & {\cellcolor[rgb]{0.545,0.647,0.839}}\textbf{99.9} & {\cellcolor[rgb]{0.847,0.882,0.949}}98.9 &  & 93.5 & 37.4 & 68.0 &  & 69.8 & 28.2 \\
 & Sky-All~$\downarrow$ & 0.016 & 0.018 &  & 0.277 & 0.988 & {\cellcolor[rgb]{0.847,0.882,0.949}}0.005 &  & {\cellcolor[rgb]{0.545,0.647,0.839}}\textbf{0.002} & 0.799 \\ 
\midrule
\multirow{3}{*}{2} & AUC~$\uparrow$ & {\cellcolor[rgb]{0.545,0.647,0.839}}\textbf{99.9} & {\cellcolor[rgb]{0.545,0.647,0.839}}\textbf{99.9} &  & {\cellcolor[rgb]{0.847,0.882,0.949}}95.9 & 65.4 & 92.3 &  & 75.6 & 58.4 \\
 & AUPR~$\uparrow$ & {\cellcolor[rgb]{0.545,0.647,0.839}}\textbf{99.8} & {\cellcolor[rgb]{0.847,0.882,0.949}}98.8 &  & 93.0 & 34.5 & 67.0 &  & 67.8 & 28.1 \\
 & Sky-All~$\downarrow$ & 0.017 & 0.018 &  & 0.280 & 0.990 & {\cellcolor[rgb]{0.847,0.882,0.949}}0.005 &  & {\cellcolor[rgb]{0.545,0.647,0.839}}\textbf{0.002} & 0.803 \\ 
\midrule
\multirow{3}{*}{3} & AUC~$\uparrow$ & {\cellcolor[rgb]{0.545,0.647,0.839}}\textbf{99.9} & {\cellcolor[rgb]{0.847,0.882,0.949}}99.7 &  & 95.9 & 62.3 & 91.6 &  & 73.6 & 58.4 \\
 & AUPR~$\uparrow$ & {\cellcolor[rgb]{0.545,0.647,0.839}}\textbf{99.7} & {\cellcolor[rgb]{0.847,0.882,0.949}}98.1 &  & 92.8 & 32.8 & 65.7 &  & 64.5 & 28.2 \\
 & Sky-All~$\downarrow$ & 0.018 & 0.020 &  & 0.283 & 0.992 & {\cellcolor[rgb]{0.847,0.882,0.949}}0.005 &  & {\cellcolor[rgb]{0.545,0.647,0.839}}\textbf{0.002} & 0.809 \\ 
\midrule
\multirow{3}{*}{4} & AUC~$\uparrow$ & {\cellcolor[rgb]{0.545,0.647,0.839}}\textbf{99.6} & {\cellcolor[rgb]{0.847,0.882,0.949}}99.5 &  & 96.1 & 58.8 & 91.8 &  & 71.3 & 58.4 \\
 & AUPR~$\uparrow$ & {\cellcolor[rgb]{0.545,0.647,0.839}}\textbf{99.1} & {\cellcolor[rgb]{0.847,0.882,0.949}}97.2 &  & 92.9 & 31.2 & 67.2 &  & 60.0 & 28.3 \\
 & Sky-All~$\downarrow$ & 0.023 & 0.022 &  & 0.288 & 0.994 & {\cellcolor[rgb]{0.847,0.882,0.949}}0.005 &  & {\cellcolor[rgb]{0.545,0.647,0.839}}\textbf{0.002} & 0.819 \\ 
\midrule
\multirow{3}{*}{5} & AUC~$\uparrow$ & {\cellcolor[rgb]{0.847,0.882,0.949}}98.5 & {\cellcolor[rgb]{0.545,0.647,0.839}}\textbf{99.0} &  & 96.5 & 58.5 & 92.2 &  & 66.8 & 57.8 \\
 & AUPR~$\uparrow$ & {\cellcolor[rgb]{0.545,0.647,0.839}}\textbf{97.1} & {\cellcolor[rgb]{0.847,0.882,0.949}}96.1 &  & 93.7 & 32.8 & 70.4 &  & 53.8 & 28.2 \\
 & Sky-All~$\downarrow$ & 0.035 & 0.026 &  & 0.295 & 0.996 & {\cellcolor[rgb]{0.847,0.882,0.949}}0.005 &  & {\cellcolor[rgb]{0.545,0.647,0.839}}\textbf{0.002} & 0.839 \\
\bottomrule
\end{tabular}
}
\caption{\textbf{Epistemic uncertainty estimation results encountering unseen pattern on Monocular depth estimation task.} The evaluation datasets used here are KITTI Seg-Depth (S$=$0) and KITTI Seg-Depth-C (S$>$0).}
\label{tab:epismde2}
\end{table}

\paragraph{4.3.3 Robustness on unseen patterns during training}
This experiment focuses on how uncertainty estimators behave on unseen patterns during training. The unseen patterns are drawn from the same dataset distribution as the patterns used in training, and the outputs of the main task model for such patterns may be reasonable. Still, they cannot be evaluated and thus are unreliable. High uncertainty should be assigned to these predictions.
Since this topic is rarely considered in MDE, we try to give a benchmark in this work.

\textbf{Evaluation settings and datasets\quad}We select sky areas in KITTI as OOD patterns. This setting is based on the following reasons: due to the generalization ability of MDE DNNs, it is inappropriate to treat all pixels without ground truth as OOD. However, there is consistently no ground truth for the sky parts since LIDAR is used in depth acquisition. During training, sky patterns are masked and never seen by the DNNs (including the AuxUEs).
Meanwhile, they are annotated in KITTI semantic segmentation dataset~\cite{Alhaija2018IJCV} (200 images), thus can be used for evaluation.

Three metrics are applied for evaluating OOD detection performance as shown in Tab.~\ref{tab:epismde2}. \textit{\textbf{AUC}} and \textit{\textbf{AUPR}}: we select 49 images that are not in the training set and have both depth and semantic segmentation annotations. For each image, we take the sky pixels as the positive class and the pixels with depth ground truth as the negative class. We use AUC and AUPR to assess the uncertainty estimation performance. Note that this metric does not guarantee that the uncertainty of the sky is the largest in the whole uncertainty map. Thus, we have \textit{\textbf{Sky-All}} (lower is better): all 200 images with semantic segmentation annotations are selected for evaluation. The ground truth uncertainties are set as $\mathbf{1}$ for the sky areas. Then we normalize the predicted uncertainty, take the sky areas $\hat{\mathbf{u}}_{\text{sky}}$ from the whole uncertainty map and measure: $mean((\mathbf{1} - \hat{\mathbf{u}}_{\text{sky}})^2)$. For simplicity, we denote KITTI Seg-Depth for both evaluation datasets. We also generate a corruption dataset KITTI Seg-Depth-C using the same way in the aleatoric uncertainty estimation section.

\textbf{Results\quad}Fig.~\ref{fig:vis_mde} shows a qualitative example of typical uncertainty maps computed on KITTI images. More visualizations are presented in Supp Section~\ref{supp:E}.
{In Tab.~\ref{tab:epismde2}, the Deep Ensembles and Gradient-based methods can better assign consistent and higher uncertainty to the sky areas, but they are 
inadequate for identifying the ID and OOD areas. 
\Xuanlong{As outlined in Section 3.2.3, DIDO prioritizes rare patterns and then generalizes the uncertainty estimation ability to the unseen patterns. This results in assigning higher uncertainty to some few-shot pixels that have ground truth, making Sky-All results slightly worse. Yet, it can achieve a balanced performance on all the metrics, and at the same time, it maintains robust performance in the presence of noise.}}

\subsection{Ablation study}
We conduct the ablation study on the corresponding section in Supp Section~\ref{supp:D}.
\noindent\textbf{Hyperparameters.}
We analyze the effect of the number of sets $K$ defined in Section~\ref{sec:discritization} for discretization and $\lambda$ for the regularization term in Eq.~\ref{eq:loss_epis}.
\noindent\textbf{Necessity of using AuxUE.}
We also apply DIDO on the main task model to check the impact on main task performance.
\noindent\textbf{Effectiveness of Dirichlet modeling.}
We show the effectiveness of the Dirichlet modeling instead of using the normal Categorical modeling based on the discretized prediction errors. For the former, we apply classical cross-entropy on the Softmax outputs given by the AuxUE.

%% file: 4supp.tex
\appendix

\renewcommand{\theequation}{A\arabic{equation}}
\renewcommand{\thetable}{A\arabic{table}}
\renewcommand{\thefigure}{A\arabic{figure}}

\newcommand*{\dictchar}[1]{
    \clearpage
    \twocolumn[
    \centerline{\parbox[c][3cm][c]{17cm}{
            \centering
            \fontsize{14}{14}
            \selectfont
            {#1}}}]
}

\dictchar{\textbf{Discretization-Induced Dirichlet Posterior for Robust Uncertainty Quantification on Regression}\vspace{0.5em}\\\textbf{------------ Supplementary Material ------------}}

\section{Supplements for the experiments}
\label{supp:A}
\subsection{Toy example: 1D signal}
\label{supp:A.1}

\paragraph{A.1.1 Dataset}
We created two toy datasets for our experiment. 
Fig.2-A on the main paper was generated as follows: $y = 10\sin(x) + \epsilon$, with $\epsilon$:
$$
\begin{cases}
    \epsilon \sim \mathcal{N}(0,3) & x\in[-3,0]\\
    \epsilon \sim \mathcal{N}(0,1) & x\in[0,3]\\
    0 & \text{otherwise}
\end{cases}
$$\noindent
Fig.2-B on the main paper was generated as follows: $y = 10\sin(x) + \epsilon$, with $\epsilon$:
$$
\begin{cases}
    \epsilon \sim \mathcal{N}(0,3) & x\in[-3,-1]\\
    \epsilon \sim \mathcal{N}(0,1) & x\in[3,5]\\
    0 & \text{otherwise}
\end{cases}
$$

\paragraph{A.1.2 Models}
Our main task model consists of an MLP with four hidden layers with 300 hidden units per layer and ReLU non-linearities. We use a generalized AuxUE method similar to ConfidNet~\cite{corbiere2019addressing}.

In particular, the input of the AuxUE is the features from the output of the penultimate layer of the main task model. Thus, in this case, the generalized AuxUE does not need the encoders, as shown in the general process in Fig.1 on the main paper. The architecture of this AuxUE is as follows. 
$\sigma_{\Theta_1}$ is composed of one fully connected layer (FCL) with an exponential activation function on the top. $\sigma_{\Theta_2}$ is composed of an MLP with a cosine similarity layer and a hidden layer with 300 hidden units per layer, and an exponential activation function on the top.

The reason for using the cosine similarity layer is to decrease the impact of the numerical value. This operation is similar to the fully connected layer operation but simply divides the output by the product of the norm of the layer's inputs (trainable parameters of the linear and input features).

\paragraph{A.1.3 Training}
The hyperparameters are listed in Tab.~\ref{tab:toy_supp_1}. As a reminder, $\lambda$ and $K$ are the hyperparameters specifically for AuxUE ($\sigma_{\Theta_2}$), which stand for the weight for the regularization term in the loss function, and respectively for the number of the class we set for discretization.
\begin{table}[t]
\centering
\scalebox{0.8}{
\begin{tabular}{lcc} 
\toprule
Hyperparameters & Main task & AuxUE \\ 
\toprule
learning rate & 0.001 & 0.005 \\
$\#$ epochs & 200 & 100 \\
batch size & 64 & 64 \\
$\lambda$ & - & 0.001 \\
$K$ & - & 5 \\
\bottomrule
\end{tabular}
}
\caption{\textbf{Hyperparameters for 1D signal toy examples.}}
\label{tab:toy_supp_1}
\end{table}

\begin{table}[t]
\centering
\scalebox{0.8}{
\begin{tabular}{lcc} 
\toprule
Hyperparameters & Main task & AuxUE \\ 
\toprule
learning rate & 0.001 & 0.001 \\
$\#$ epochs & 150 & 20 \\
batch size & 64 & 64 \\
$\lambda$ & - & 0.0001 \\
$K$ & - & 5 \\
\bottomrule
\end{tabular}
}
\caption{\textbf{Hyperparameters for tabular data example.}}
\label{tab:toy_supp_2}
\end{table}

\subsection{Tabular data example}\label{supp:A2}
\paragraph{A.2.1 Dataset}
In the tabular data example, we use the red wine quality dataset~\cite{cortez2009modeling} for the OOD detection task. We randomly separate the dataset in training, validation, and test sets with 72\%, 8\%, and 20\% as the proportions of the whole dataset for each set. We generate the OOD data using the ID test set. We first replicate two test sets as OOD sets, one of which we set all the features in the table to be negative, and the other, we randomly shuffle the values of the features.

\paragraph{A.2.2 Models and training}
Our main task model consists of an MLP with four hidden layers with 16, 32, and 16 hidden units in the respective layer and ReLU non-linearities. We use a generalized AuxUE method
similar to ConfidNet~\cite{corbiere2019addressing}.

In particular, we find it better to provide the tabular data to the AuxUE directly. We use one hidden layer with 16 hidden units followed by ReLU as the feature extractor for $\sigma_{\Theta_1}$ and $\sigma_{\Theta_2}$ uncertainty estimators. For the uncertainty estimators, we use the same ones as in the 1D signal data. 
The hyperparameters are listed in Tab.~\ref{tab:toy_supp_2}.

\paragraph{A.2.3 Results}
We trained three models to build Deep Ensembles (DEns.)~\cite{lakshminarayanan2017simple}. The epistemic uncertainty estimates are obtained using the variance of DNNs' point estimates. We evaluate the OOD detection performance using AUC and AUPR as the metrics. The results are shown in Tab.~\ref{tab:toy_supp_result_2}. The proposed DIDO outperforms the DEns. on OOD detection task using one extra DNN apart from the main task model.

\begin{table}[t]
\centering
\scalebox{0.8}{
\begin{tabular}{lccc} 
\toprule
 & MSE $\downarrow$ & AUC $\uparrow$ & AUPR $\uparrow$ \\ 
\bottomrule
DEns. &\first \textbf{0.646} & 0.548 & 0.250 \\
DIDO &\first \textbf{0.646} &\first \textbf{0.936} &\first \textbf{0.863} \\
\bottomrule
\end{tabular}}
\caption{\textbf{Main task and OOD detection performance on tabular data example.}}
\label{tab:toy_supp_result_2}
\end{table}

\subsection{Age estimation}\label{supp:A3}
\paragraph{A.3.1 Model}
The main task ResNet34~\cite{he2016deep} model checkpoints are downloaded from the official GitHub repository of Coral~\cite{cao2020rank}. We observe that the age estimation result can outperform the one achieved by Coral by applying soft-weighted-sum (SWS)~\cite{yu_car} on the top of the models trained using cross-entropy loss. The goal of SWS is a post-processing operation to transfer the discrete Softmax outputs to the continuous age estimates. For this reason, we use the main task models trained by cross-entropy loss.  

As introduced in Section 4.2 of the main paper, the AuxUE DNN is applied in a ConfidNet~\cite{corbiere2019addressing} way. Similarly to the toy example settings, we take the pre-logits (512 features) from the main task model as the inputs of our AuxUE.

For $\sigma_{{\Theta}_1}$, we use an MLP with one hidden layer with 512 hidden units and an FCL with an exponential function on the top.
For $\sigma_{{\Theta}_2}$, we use an MLP with a cosine similarity layer and one hidden layer with 512 hidden units per layer and ReLU non-linearities, followed by an FCL with an exponential function on the top.

\paragraph{A.3.2 Training}
To train the AuxUE DNN, we use the hyperparameters shown in Tab.~\ref{tab:age_supp}. We use the same optimizer and batch size as for the main task training, while we use 25 epochs which is much less than training the main task.

\begin{table}
\centering
\scalebox{0.8}{
\begin{tabular}{lcc} 
\toprule
Hyperparameters & Main task & AuxUE \\ 
\toprule
learning rate & 0.0005 & 0.001 \\
$\#$ epochs & 200 & 25 \\
batch size & 256 & 256 \\
$\lambda$ & - & 0.01 \\
$K$ & - & 8 \\
\bottomrule
\end{tabular}
}
\caption{\textbf{Hyperparameters for age estimation.}}
\label{tab:age_supp}
\end{table}

\paragraph{A.3.3 Main task and aleatoric uncertainty performance}
\label{sec:age_supp}
For the age estimation task, we list the main task results in Tab.~\ref{tab:mainresult_ae_supp} given by the original Coral, the original cross entropy (CE)-based models and the CE-based models using soft-weighted-sum (SWS). We can see that SWS really improves the main task performance.  Furthermore, by adjusting the original model to output the parameters of Gaussian distribution~\cite{nix1994uncertainty,kendall2017uncertainties} and training three models like this from scratch, we can achieve the results given by Deep Ensembles (DEns.)~\cite{lakshminarayanan2017simple}. We also implement LDU~\cite{franchi2022latent} and Evidential learning (Evi.)~\cite{joo2020being} based on the ResNet34 backbone. The overall difference among different techniques is not huge, while the adjustments still reduce a bit the age estimation performance. We argue that the adjusted DNNs might achieve comparable performance to the unchanged ones, but more tuning and hyperparameter searching should be required. On the other hand, for aleatoric uncertainty estimation result, for AUSE-RMSE ($\downarrow$), Ours: 0.067, LDU: 0.056, Evi.: 0.070, DEns.: 0.074, Grad.: 0.073, Inject.: 0.076. Ours is shown to provide comparable results to the other solutions.

\begin{table}
\centering
\scalebox{0.65}{
\begin{tabular}{lcccccc} 
\toprule
Metrics & Coral & CE & \begin{tabular}[c]{@{}c@{}}CE\\+SWS\end{tabular} & LDU & Evi. & DEns. \\ 
\toprule
MAE~$\downarrow$ & 3.47 $\pm$ 0.05 & 3.60 $\pm$ 0.02 & \second 3.39 & 3.41 & 3.70 $\pm$ 0.19 & \first\first \textbf{3.31} \\
RMSE~$\downarrow$ & 4.71 $\pm$ 0.06 & 5.03 $\pm$ 0.03 & 4.52 $\pm$ 0.03 & \second 4.50 & 4.72 $\pm$ 0.23 & \first\first \textbf{4.40} \\
\bottomrule
\end{tabular}
}
\caption{\textbf{Main task performance for ResNet34 model based on different methods on age estimation task.} The evaluation is based on AFAD~\cite{Niu_2016_CVPR} test set.}
\label{tab:mainresult_ae_supp}
\end{table}

\subsection{Super-resolution}\label{supp:A4}
In the SR task, the noise in the reconstructed image will be irreducible given the noisy low-resolution input, and we consider this uncertainty to be aleatoric. Moreover, we argue that the definition of epistemic uncertainty is rather vague in this task. Therefore, in this section, we use AuxUE to estimate the aleatoric uncertainty based on different distribution assumptions.
\paragraph{A.4.1 Model}
Similar to the monocular depth estimation experiments in Section 4.3.1 in the main paper, we choose SRGan~\cite{ledig2017photo} as the main task model and BayesCap~\cite{upadhyay2022bayescap} as the AuxUE and follow the same training and evaluation settings as in~\cite{upadhyay2022bayescap}.
The goal is to analyze the fundamental performance and robustness of aleatoric uncertainty estimation under different distribution assumptions. We choose simple Gaussian (Sgau)~\cite{nix1994uncertainty}, Laplacian (Lap), Generalized Gaussian (Ggau)~\cite{upadhyay2022bayescap} and Normal-Inverse-Gamma (NIG)~\cite{amini2020deep} distributions on BayesCap~\cite{upadhyay2022bayescap}. 

We modify the loss functions to output the desired parameters of the distributions.
For the architecture adjustments, we only modify the prediction heads on Bayescap. Original BayesCap~\cite{upadhyay2022bayescap} uses multiple Residual blocks~\cite{he2016deep} followed by three heads which output the three parameters for the Generalized Gaussian distribution, including one as the refined main task prediction. Each head contains a set of convolutional layers + PReLU activation functions. As we apply different distribution assumptions, we use the different numbers of the same heads to construct the variants of BayesCap. Specifically, we use two heads for two Gaussian distribution parameters, two heads for two Laplace distribution parameters, and four heads for four parameters in NIG distribution.

\paragraph{A.4.2 Training}
We follow the same training settings (batch size, learning rate, weight for the additional identity mapping loss, and the number of epochs) as in the original paper~\cite{upadhyay2022bayescap}.

\paragraph{A.4.3 Evaluation settings and datasets}
We follow~\cite{upadhyay2022bayescap} to use the Uncertainty Calibration Error (\textbf{\textit{UCE}}, lower is better) metric~\cite{laves2020calibration}. It measures the difference between the predicted uncertainty and the prediction error. Specifically, the prediction error and estimated uncertainty are assigned into bins, and the absolute difference between the mean prediction error and mean estimated uncertainty in each bin is calculated. UCE is the sum of the results from all bins.

We use ImageNet~\cite{deng2009imagenet} as the training set for both SRGan and BayesCap models. For uncertainty evaluation, we use Set5~\cite{bevilacqua2012low}, Set14~\cite{zeyde2012single}, and BSDS100~\cite{martin2001database} as the testing sets. Moreover, we generate Set5-C, Set14-C, and BSDS100-C using the code of ImageNet-C~\cite{hendrycks2019benchmarking} to have different corruptions on the images. We apply the following eighteen perturbations with five severities: Gaussian noise, shot noise, impulse noise, iso noise, defocus blur, glass blur, motion blur, zoom blur, frost, fog, snow, dark, brightness, contrast, pixelated, elastic, color quantization, and JPEG. In the main paper, we mentioned these perturbations in Section 4.3, yet due to the paper limitation, we put the complete list here.
Only low-resolution images (inputs) are polluted by noise, while the corresponding high-resolution ground truth images are clean. Castillo et al.~\cite{castillo2021generalized} applied the noise to the input images during training, while we apply them during inference for robust uncertainty estimation evaluation.

\paragraph{A.4.4 Results}
As shown in Tab.~\ref{tab:alea_sr_dido}, the Laplacian assumption on the data-dependent noise performs better than all the other assumptions, including the Generalized Gaussian distribution proposed in BayesCap.
When the noise severity increases, using the Laplacian assumption can provide more robust uncertainty than the others.

\begin{table}[t]
\centering
\arrayrulecolor{black}
\scalebox{0.8}{
\begin{tabular}{lccccc}
\multicolumn{6}{l}{\textbf{Super Resolution (Metric:~\textbf{UCE}} $\downarrow$)} \\ 
\toprule
Dataset & S & \begin{tabular}[c]{@{}c@{}}Original\\(Ggau)\end{tabular} & + Sgau & + NIG & \begin{tabular}[c]{@{}c@{}}Ours $\sigma_{\Theta_1}$\\(+ Lap)\end{tabular} \\ 
\toprule
\multirow{6}{*}{Set5} & 0 & 0.0088 & 0.0083 & {\cellcolor[rgb]{0.557,0.663,0.859}}\textbf{0.0018} & {\cellcolor[rgb]{0.851,0.882,0.949}}0.0019 \\
 & 1 & 0.0186 & 0.0180 & {\cellcolor[rgb]{0.557,0.663,0.859}}\textbf{0.0156} & {\cellcolor[rgb]{0.851,0.882,0.949}}0.0157 \\
 & 2 & 0.0253 & 0.0243 & {\cellcolor[rgb]{0.557,0.663,0.859}}\textbf{0.0226} & {\cellcolor[rgb]{0.851,0.882,0.949}}0.0227 \\
 & 3 & 0.0363 & 0.0341 & {\cellcolor[rgb]{0.851,0.882,0.949}}0.0333 & {\cellcolor[rgb]{0.557,0.663,0.859}}\textbf{\textbf{0.0332}} \\
 & 4 & 0.0434 & 0.0394 & {\cellcolor[rgb]{0.851,0.882,0.949}}0.0392 & {\cellcolor[rgb]{0.557,0.663,0.859}}\textbf{\textbf{0.0389}} \\
 & 5 & 0.0525 & 0.0462 & 0.0464 & {\cellcolor[rgb]{0.557,0.663,0.859}}\textbf{0.0040} \\ 
\midrule
\multirow{6}{*}{Set14} & 0 & 0.0137 & 0.0092 & {\cellcolor[rgb]{0.557,0.663,0.859}}\textbf{0.0040} & {\cellcolor[rgb]{0.557,0.663,0.859}}\textbf{0.0040} \\
 & 1 & 0.0221 & 0.0195 & {\cellcolor[rgb]{0.851,0.882,0.949}}0.0176 & {\cellcolor[rgb]{0.557,0.663,0.859}}\textbf{0.0174} \\
 & 2 & 0.0281 & 0.0255 & {\cellcolor[rgb]{0.851,0.882,0.949}}0.0241 & {\cellcolor[rgb]{0.557,0.663,0.859}}\textbf{0.0240} \\
 & 3 & 0.0350 & 0.0318 & {\cellcolor[rgb]{0.851,0.882,0.949}}0.0310 & {\cellcolor[rgb]{0.557,0.663,0.859}}\textbf{0.0308} \\
 & 4 & 0.0408 & 0.0368 & {\cellcolor[rgb]{0.851,0.882,0.949}}0.0364 & {\cellcolor[rgb]{0.557,0.663,0.859}}\textbf{0.0362} \\
 & 5 & 0.0509 & {\cellcolor[rgb]{0.851,0.882,0.949}}0.0465 & {\cellcolor[rgb]{0.851,0.882,0.949}}0.0465 & {\cellcolor[rgb]{0.557,0.663,0.859}}\textbf{0.0461} \\ 
\midrule
\multirow{6}{*}{BSDS100} & 0 & 0.0124 & 0.0071 & {\cellcolor[rgb]{0.851,0.882,0.949}}0.0036 & {\cellcolor[rgb]{0.557,0.663,0.859}}\textbf{\textbf{0.0033}} \\
 & 1 & 0.0204 & 0.0174 & {\cellcolor[rgb]{0.851,0.882,0.949}}0.0162 & {\cellcolor[rgb]{0.557,0.663,0.859}}\textbf{\textbf{0.0160}} \\
 & 2 & 0.0271 & 0.0237 & {\cellcolor[rgb]{0.851,0.882,0.949}}0.0229 & {\cellcolor[rgb]{0.557,0.663,0.859}}\textbf{0.0227} \\
 & 3 & 0.0332 & 0.0288 & {\cellcolor[rgb]{0.851,0.882,0.949}}0.0286 & {\cellcolor[rgb]{0.557,0.663,0.859}}\textbf{\textbf{0.0358}} \\
 & 4 & 0.0425 & 0.0363 & {\cellcolor[rgb]{0.851,0.882,0.949}}0.0363 & {\cellcolor[rgb]{0.557,0.663,0.859}}\textbf{\textbf{0.0358}} \\
 & 5 & 0.0539 & {\cellcolor[rgb]{0.851,0.882,0.949}}0.0459 & 0.0460 & {\cellcolor[rgb]{0.557,0.663,0.859}}\textbf{0.0453} \\
\bottomrule
\end{tabular}
}
\caption{\textbf{Aleatoric uncertainty estimation results on Super-Resolution task.} 
Datasets with an S (severity) greater than 1 are the -C
variants of the corresponding clean dataset.}
\label{tab:alea_sr_dido}
\end{table}

\paragraph{A.4.5 Main task performance}
In the super-resolution task, we take the main task SRGan~\cite{ledig2017photo} model used in BayesCap~\cite{upadhyay2022bayescap}. Thus we have the same main task performance as they showed in the paper. We list the results in Tab.~\ref{tab:mainresult_sr_supp} as a reminder. The evaluation is based on Set5~\cite{bevilacqua2012low}, Set14~\cite{zeyde2012single}, BSDS100~\cite{martin2001database} dataset.

\begin{table}
\centering
\scalebox{0.8}{
\centering
\begin{tabular}{cccc} \toprule
Metrics & Set5 & Set14 & BSDS100 \\ \toprule
PSNR~$\uparrow$  & 29.40 & 26.02 & 25.16 \\
SSIM~$\uparrow$  & 0.8472 & 0.7397 & 0.6688 \\ \bottomrule
\end{tabular}
}
\caption{\textbf{Main task performance for SRGan model on super-resolution task.}}
\label{tab:mainresult_sr_supp}
\end{table}

\subsection{Monocular depth estimation}
\paragraph{A.5.1 Model}
We use SLURP~\cite{yu2021slurp} as the backbone in this experiment. We modify the prediction heads to achieve the uncertainty estimates.

For $\sigma_{\Theta_1}$, we do not modify the model when the distribution assumption only contains one parameter (except for the main task prediction term).
For the distribution assumptions with more than one parameter output, we add one more convolutional layer with ReLU on the top for a fair comparison.

For $\sigma_{\Theta_2}$, similarly to the one in age estimation, we replace the original head (a single convolutional layer) with the cosine similarity layer followed by two convolutional layers with ReLU activation functions. 
In the cases where we share the encoders to make the general AuxUE lighter, based on the original SLURP, we doubled the number of features fed into the prediction head. We split them into two sets, feeding them into two prediction heads. The two prediction heads are consistent in the structures mentioned before.
We follow~\cite{yu2021slurp} to use the depth output and the encoder features of the main task model BTS~\cite{lee2019big} as the input of the AuxUE.

\paragraph{A.5.2 Training}
The hyperparameters used during training are listed in Tab.~\ref{tab:mde_supp}. The learning rate decrement is consistent with the main task BTS~\cite{lee2019big} model.
\begin{table}
\centering
\scalebox{0.8}{
\begin{tabular}{lcc} 
\toprule
Hyperparameters & Main task & AuxUE \\ 
\toprule
start learning rate & 1e-4 & 1e-4 \\
end learning rate & 1e-5 & 1e-5 \\
$\#$ epochs & 50 & 8 \\
batch size & 4 & 4 \\
$\lambda$ & - & 0.01 \\
$K$ & - & 32 \\
\bottomrule
\end{tabular}
}
\caption{\textbf{Hyperparameters for monocular depth estimation.}}
\label{tab:mde_supp}
\end{table}

\paragraph{A.5.3 Main task performance}
In monocular depth estimation, we list in Tab.~\ref{tab:mainresult_mde_supp} the results for the methods using modified main task BTS~\cite{lee2019big} models, namely SinglePU~\cite{kendall2017uncertainties}, Deep Ensembles (DEns.)~\cite{lakshminarayanan2017simple}, LDU~\cite{franchi2022latent}, as well as the original model, which is used for AuxUEs and the training-free methods. Note that we use the evaluation code based on AdaBins~\cite{bhat2021adabins}, which corrected the error made in the BTS evaluation code, and the result will be slightly better than the one claimed in the original BTS.  The evaluation is based on KITTI~\cite{Geiger2013IJRR} Eigen-split~\cite{eigen2014depth} validation set. As we can see, modifying the model and training in~\cite{kendall2017uncertainties} way will affect the main task performance even after doing Deep Ensembles, LDU can provide competitive results to the original yet only on several metrics. Overall, the AuxUE is necessary to be applied for uncertainty estimation without changing and affecting the main task.

\begin{table}
\centering
\scalebox{0.65}{
\begin{tabular}{lcccccccc} \toprule
Methods & absrel~$\downarrow$ & log10~$\downarrow$ & rms~$\downarrow$ & sqrel~$\downarrow$ & logrms~$\downarrow$ & d1~$\uparrow$ & d2~$\uparrow$ & d3~$\uparrow$ \\ \toprule
Org &\first \textbf{0.056} &\first \textbf{0.025} & {2.430} &\first \textbf{0.201} &\first \textbf{0.089} &\first \textbf{0.963} & {0.994} &\first \textbf{0.999} \\
SinglePU & 0.065 & 0.029 & 2.606 & 0.234 & 0.100 & 0.952 & 0.993 & {0.998} \\
LDU & {0.059} & {0.026} &\first \textbf{2.394} & 0.203 & {0.091} & 0.960 & {0.994} &\first \textbf{0.999} \\
DEns. & 0.060 & {0.026} & 2.435 & {0.202} & 0.092 & {0.961} &\first \textbf{0.995} &\first \textbf{0.999} \\ \bottomrule
\end{tabular}
}
\caption{\textbf{Main task performance for original and modified BTS models on monocular depth estimation.} The evaluation is based on KITTI~\cite{Geiger2013IJRR} Eigen-split~\cite{eigen2014depth} validation set.}
\label{tab:mainresult_mde_supp}
\end{table}

\paragraph{A.5.4 Additional results on aleatoric uncertainty estimation from Dirichlet outputs}
The class corresponds to the maximum output value of the Dirichlet output, i.e., outputs of $\sigma_{\Theta_2}$ can also be regarded as aleatoric uncertainty estimation. We provide the corresponding results in Tab.~\ref{tab:alea_mde_supp}. We also put the partial results from the other distribution assumptions to make a comparison. From the last column, we can see that the aleatoric uncertainty given by the Dirichlet outputs underperforms the other solutions on quantitative metrics since the discretization affects the original numerical values of the prediction errors on the ID training data. Yet, the performance reduction is not huge and unacceptable.

\begin{table}
\centering
\scalebox{0.63}{
\begin{tabular}{clccccc} 
\toprule
S & \multicolumn{1}{c}{Metrics} & + Sgau & + NIG & \begin{tabular}[c]{@{}c@{}}Ours~(+ Lap)\\shar. enc. $\sigma_{\Theta_1}$\end{tabular} & \begin{tabular}[c]{@{}c@{}}Ours~(+ Lap)\\sep. enc. $\sigma_{\Theta_1}$\end{tabular} & \begin{tabular}[c]{@{}c@{}}Ours~(DIDO)\\sep. enc. $\sigma_{\Theta_2}$\end{tabular} \\ 
\toprule
\multirow{4}{*}{0} & AUSE-REL $\downarrow$ & {\cellcolor[rgb]{0.851,0.882,0.949}}{0.013} & {\cellcolor[rgb]{0.557,0.663,0.859}}\textbf{0.012} & {\cellcolor[rgb]{0.851,0.882,0.949}}{0.013} & {\cellcolor[rgb]{0.851,0.882,0.949}}{0.013} & 0.015 \\
 & AUSE-RMSE $\downarrow$ & {\cellcolor[rgb]{0.557,0.663,0.859}}\textbf{0.202} & 0.208 & 0.205 & {\cellcolor[rgb]{0.851,0.882,0.949}}{0.203} & 0.283 \\
 & AURG-REL $\uparrow$ & {\cellcolor[rgb]{0.851,0.882,0.949}}{0.023} & {\cellcolor[rgb]{0.557,0.663,0.859}}\textbf{0.024} & {\cellcolor[rgb]{0.851,0.882,0.949}}{0.023} & {\cellcolor[rgb]{0.851,0.882,0.949}}{0.023} & 0.021 \\
 & AURG-RMSE $\uparrow$ & {\cellcolor[rgb]{0.557,0.663,0.859}}\textbf{1.871} & 1.865 & 1.869 & {\cellcolor[rgb]{0.851,0.882,0.949}}{1.870} & 1.791 \\ 
\midrule
\multirow{4}{*}{1} & AUSE-REL $\downarrow$ & {\cellcolor[rgb]{0.851,0.882,0.949}}{0.019} & {\cellcolor[rgb]{0.557,0.663,0.859}}\textbf{0.018} & {\cellcolor[rgb]{0.557,0.663,0.859}}\textbf{0.018} & {\cellcolor[rgb]{0.851,0.882,0.949}}{0.019} & 0.023 \\
 & AUSE-RMSE $\downarrow$ & {\cellcolor[rgb]{0.557,0.663,0.859}}\textbf{0.332} & {\cellcolor[rgb]{0.851,0.882,0.949}}{0.335} & {\cellcolor[rgb]{0.557,0.663,0.859}}\textbf{0.332} & 0.336 & 0.474 \\
 & AURG-REL $\uparrow$ & {\cellcolor[rgb]{0.851,0.882,0.949}}{0.031} & {\cellcolor[rgb]{0.557,0.663,0.859}}\textbf{0.032} & {\cellcolor[rgb]{0.557,0.663,0.859}}\textbf{0.032} & {\cellcolor[rgb]{0.851,0.882,0.949}}{0.031} & 0.027 \\
 & AURG-RMSE $\uparrow$ & {\cellcolor[rgb]{0.557,0.663,0.859}}\textbf{2.365} & {\cellcolor[rgb]{0.851,0.882,0.949}}2.362 & {\cellcolor[rgb]{0.557,0.663,0.859}}\textbf{2.365} & 2.361 & 2.223 \\ 
\midrule
\multirow{4}{*}{2} & AUSE-REL $\downarrow$ & {\cellcolor[rgb]{0.851,0.882,0.949}}{0.023} & {\cellcolor[rgb]{0.557,0.663,0.859}}\textbf{0.022} & {\cellcolor[rgb]{0.557,0.663,0.859}}\textbf{0.022} & {\cellcolor[rgb]{0.851,0.882,0.949}}{0.023} & 0.028 \\
 & AUSE-RMSE $\downarrow$ & {\cellcolor[rgb]{0.557,0.663,0.859}}\textbf{0.463} & 0.479 & {\cellcolor[rgb]{0.851,0.882,0.949}}{0.464} & 0.468 & 0.669 \\
 & AURG-REL $\uparrow$ & {\cellcolor[rgb]{0.557,0.663,0.859}}\textbf{0.039} & {\cellcolor[rgb]{0.557,0.663,0.859}}\textbf{0.039} & {\cellcolor[rgb]{0.557,0.663,0.859}}\textbf{0.039} & {\cellcolor[rgb]{0.851,0.882,0.949}}{0.038} & 0.033 \\
 & AURG-RMSE $\uparrow$ & {\cellcolor[rgb]{0.557,0.663,0.859}}\textbf{2.779} & 2.763 & {\cellcolor[rgb]{0.851,0.882,0.949}}{2.777} & 2.774 & 2.573 \\ 
\midrule
\multirow{4}{*}{3} & AUSE-REL $\downarrow$ & {\cellcolor[rgb]{0.557,0.663,0.859}}\textbf{0.031} & {\cellcolor[rgb]{0.557,0.663,0.859}}\textbf{0.031} & {\cellcolor[rgb]{0.557,0.663,0.859}}\textbf{0.031} & {\cellcolor[rgb]{0.557,0.663,0.859}}\textbf{\textbf{0.031}} & 0.040 \\
 & AUSE-RMSE $\downarrow$ & {\cellcolor[rgb]{0.851,0.882,0.949}}{0.737} & 0.806 & 0.749 & {\cellcolor[rgb]{0.557,0.663,0.859}}\textbf{\textbf{0.730}} & 1.093 \\
 & AURG-REL $\uparrow$ & {\cellcolor[rgb]{0.557,0.663,0.859}}\textbf{0.049} & {\cellcolor[rgb]{0.557,0.663,0.859}}\textbf{0.049} & {\cellcolor[rgb]{0.557,0.663,0.859}}\textbf{0.049} & {\cellcolor[rgb]{0.557,0.663,0.859}}\textbf{\textbf{0.049}} & 0.040 \\
 & AURG-RMSE $\uparrow$ & {\cellcolor[rgb]{0.851,0.882,0.949}}{3.301} & 3.232 & 3.289 & {\cellcolor[rgb]{0.557,0.663,0.859}}\textbf{\textbf{3.308}} & 2.945 \\ 
\midrule
\multirow{4}{*}{4} & AUSE-REL $\downarrow$ & {\cellcolor[rgb]{0.851,0.882,0.949}}{0.050} & 0.053 & 0.051 & {\cellcolor[rgb]{0.557,0.663,0.859}}\textbf{\textbf{0.049}} & 0.064 \\
 & AUSE-RMSE $\downarrow$ & {\cellcolor[rgb]{0.851,0.882,0.949}}{1.364} & 1.582 & 1.430 & {\cellcolor[rgb]{0.557,0.663,0.859}}\textbf{\textbf{1.268}} & 2.029 \\
 & AURG-REL $\uparrow$ & {\cellcolor[rgb]{0.851,0.882,0.949}}{0.058} & 0.054 & 0.056 & {\cellcolor[rgb]{0.557,0.663,0.859}}\textbf{\textbf{0.059}} & 0.044 \\
 & AURG-RMSE $\uparrow$ & {\cellcolor[rgb]{0.851,0.882,0.949}}{3.834} & 3.615 & 3.767 & {\cellcolor[rgb]{0.557,0.663,0.859}}\textbf{\textbf{3.929}} & 3.168 \\ 
\midrule
\multirow{4}{*}{5} & AUSE-REL $\downarrow$ & {\cellcolor[rgb]{0.851,0.882,0.949}}{0.064} & 0.069 & 0.066 & {\cellcolor[rgb]{0.557,0.663,0.859}}\textbf{\textbf{0.059}} & 0.080 \\
 & AUSE-RMSE $\downarrow$ & {\cellcolor[rgb]{0.851,0.882,0.949}}{2.043} & 2.414 & 2.157 & {\cellcolor[rgb]{0.557,0.663,0.859}}\textbf{\textbf{1.760}} & 3.021 \\
 & AURG-REL $\uparrow$ & {\cellcolor[rgb]{0.851,0.882,0.949}}{0.063} & 0.057 & 0.061 & {\cellcolor[rgb]{0.557,0.663,0.859}}\textbf{0.067} & 0.046 \\
 & AURG-RMSE $\uparrow$ & {\cellcolor[rgb]{0.851,0.882,0.949}}{4.213} & 3.842 & 4.098 & {\cellcolor[rgb]{0.557,0.663,0.859}}\textbf{4.496} & 3.235 \\
\bottomrule
\end{tabular}
}
\caption{\textbf{Aleatoric uncertainty estimation results on Monocular Depth Estimation task.} The additional aleatoric uncertainty estimation results are listed in the last column.}
\label{tab:alea_mde_supp}
\end{table}

\paragraph{A.5.5 Full results on epistemic uncertainty estimation}
Tab.~\ref{tab:epismde1_supp} shows the results on the robustness of different methods for epistemic uncertainty estimation under dataset change. Tab.~\ref{tab:epismde2_supp} provides the results on evaluating the robustness of the uncertainty estimation methods on unseen patterns during training.
As we can see, in Tab.~\ref{tab:epismde2_supp}, most distribution assumptions can help AuxUE achieve good AUC and AUPR results, which shows that these AuxUEs all fit the ID data well. Yet, they can not assign consistent and higher uncertainty to the sky areas.
\begin{table*}[t]
\centering
\scalebox{0.72}{
\begin{tabular}{lccccccccccccccc} 
\toprule
 & \multicolumn{7}{c}{Auxiliary Uncertainty Estimator (SLURP)~} &  & \multicolumn{4}{c}{Modified task DNN} &  & \multicolumn{2}{c}{Training-free} \\ 
\cmidrule[\heavyrulewidth]{2-8}\cmidrule[\heavyrulewidth]{10-13}\cmidrule[\heavyrulewidth]{15-16}
Metrics & Original & + Ggau & + Sgau & + NIG & \begin{tabular}[c]{@{}c@{}}Ours $\sigma_{\Theta_1}$sep.\\enc.(+Lap)\end{tabular} & \begin{tabular}[c]{@{}c@{}}Ours $\sigma_{\Theta_2}$sep.\\enc.(DIDO)\end{tabular} & \begin{tabular}[c]{@{}c@{}}Ours $\sigma_{\Theta_2}$shar.\\enc.(DIDO)\end{tabular} &  & \begin{tabular}[c]{@{}c@{}}Single\\PU\end{tabular} & LDU & Evi. & DEns. &  & Grad. & Inject. \\ 
\toprule
AUC~$\uparrow$ & 59.8 & 80.9 & 74.5 & 57.0 & 65.4 & {\cellcolor[rgb]{0.843,0.882,0.949}}98.1 & {\cellcolor[rgb]{0.557,0.667,0.863}}\textbf{98.4} &  & 64.2 & 58.1 & 70.6 & 62.1 &  & 78.4 & 18.3 \\
AUPR~$\uparrow$ & 76.7 & 90.9 & 88.4 & 75.5 & 82.5 & {\cellcolor[rgb]{0.843,0.882,0.949}}99.3 & {\cellcolor[rgb]{0.557,0.667,0.863}}\textbf{99.4} &  & 78.3 & 79.5 & 77.8 & 76.7 &  & 92.6 & 62.3 \\
\bottomrule
\end{tabular}
}
\caption{\textbf{Epistemic uncertainty estimation results encountering dataset change on Monocular depth estimation task.} The evaluation dataset used here is NYU indoor depth dataset.}
\label{tab:epismde1_supp}
\end{table*}

\begin{table*}[t]
\centering
\scalebox{0.72}{
\begin{tabular}{clccccccccccccccc} 
\toprule
 &  & \multicolumn{7}{c}{Auxiliary Uncertainty Estimator (SLURP)~} &  & \multicolumn{4}{c}{Modified main DNN} &  & \multicolumn{2}{c}{Training-free} \\ 
\cmidrule[\heavyrulewidth]{3-9}\cmidrule[\heavyrulewidth]{11-14}\cmidrule[\heavyrulewidth]{16-17}
S & Metrics & Original & + Ggau & + Sgau & + NIG & \begin{tabular}[c]{@{}c@{}}Ours $\sigma_{\Theta_1}$ sep.\\enc.(+Lap)\end{tabular} & \begin{tabular}[c]{@{}c@{}}Ours $\sigma_{\Theta_2}$ sep.\\enc.(DIDO)\end{tabular} & \begin{tabular}[c]{@{}c@{}}Ours $\sigma_{\Theta_2}$ shar.\\enc.(DIDO)\end{tabular} &  & \begin{tabular}[c]{@{}c@{}}Single\\PU\end{tabular} & LDU & Evi. & DEns. &  & \begin{tabular}[c]{@{}c@{}}Grad.\\(inv.)\end{tabular} & Inject. \\ 
\toprule
\multirow{3}{*}{0} & AUC~$\uparrow$ & 99.1 & 96.8 & 99.0 & 90.9 & {\cellcolor[rgb]{0.847,0.882,0.949}}99.9 & {\cellcolor[rgb]{0.545,0.647,0.839}}\textbf{100.0} & {\cellcolor[rgb]{0.847,0.882,0.949}}99.9 &  & 89.0 & 96.5 & 76.7 & 93.5 &  & 85.6 & 58.4 \\
 & AUPR~$\uparrow$ & 94.6 & 80.3 & 91.6 & 57.6 & {\cellcolor[rgb]{0.847,0.882,0.949}}99.7 & {\cellcolor[rgb]{0.545,0.647,0.839}}\textbf{100.0} & 99.0 &  & 62.0 & 93.8 & 42.6 & 70.0 &  & 76.3 & 28.1 \\
 & Sky-All~$\downarrow$ & 0.643 & 0.277 & 0.934 & 0.983 & 0.961 & 0.015 & 0.018 &  & {\cellcolor[rgb]{0.847,0.882,0.949}}0.005 & 0.278 & 0.986 & {\cellcolor[rgb]{0.847,0.882,0.949}}0.005 &  & {\cellcolor[rgb]{0.545,0.647,0.839}}\textbf{0.001} & 0.800 \\ 
\midrule
\multirow{3}{*}{1} & AUC~$\uparrow$ & 98.4 & 96.0 & 99.0 & 90.4 & 99.8 & {\cellcolor[rgb]{0.545,0.647,0.839}}\textbf{100.0} & {\cellcolor[rgb]{0.847,0.882,0.949}}99.9 &  & 86.9 & 96.3 & 69.7 & 92.8 &  & 76.9 & 58.5 \\
 & AUPR~$\uparrow$ & 93.3 & 77.9 & 92.4 & 57.4 & {\cellcolor[rgb]{0.847,0.882,0.949}}99.5 & {\cellcolor[rgb]{0.545,0.647,0.839}}\textbf{99.9} & 98.9 &  & 59.1 & 93.5 & 37.4 & 68.0 &  & 69.8 & 28.2 \\
 & Sky-All~$\downarrow$ & 0.742 & 0.274 & 0.935 & 0.978 & 0.962 & 0.016 & 0.018 &  & {\cellcolor[rgb]{0.847,0.882,0.949}}0.005 & 0.277 & 0.988 & {\cellcolor[rgb]{0.847,0.882,0.949}}0.005 &  & {\cellcolor[rgb]{0.545,0.647,0.839}}\textbf{0.002} & 0.799 \\ 
\midrule
\multirow{3}{*}{2} & AUC~$\uparrow$ & 97.6 & 95.6 & 99.0 & 90.2 & {\cellcolor[rgb]{0.847,0.882,0.949}}99.7 & {\cellcolor[rgb]{0.545,0.647,0.839}}\textbf{99.9} & {\cellcolor[rgb]{0.545,0.647,0.839}}\textbf{99.9} &  & 86.6 & 95.9 & 65.4 & 92.3 &  & 75.6 & 58.4 \\
 & AUPR$\uparrow$ & 91.9 & 76.5 & 92.8 & 57.5 & {\cellcolor[rgb]{0.847,0.882,0.949}}99.3 & {\cellcolor[rgb]{0.545,0.647,0.839}}\textbf{99.8} & 98.8 &  & 58.9 & 93.0 & 34.5 & 67.0 &  & 67.8 & 28.1 \\
 & Sky-All~$\downarrow$ & 0.784~ & 0.274 & 0.937 & 0.973 & 0.962 & 0.017 & 0.018 &  & {\cellcolor[rgb]{0.847,0.882,0.949}}0.005 & 0.280 & 0.990 & {\cellcolor[rgb]{0.847,0.882,0.949}}0.005 &  & {\cellcolor[rgb]{0.545,0.647,0.839}}\textbf{0.002} & 0.803 \\ 
\midrule
\multirow{3}{*}{3} & AUC~$\uparrow$ & 96.8 & 95.0 & 98.9 & 90.0 & 99.4 & {\cellcolor[rgb]{0.545,0.647,0.839}}\textbf{99.9} & {\cellcolor[rgb]{0.847,0.882,0.949}}99.7 &  & 86.6 & 95.9 & 62.3 & 91.6 &  & 73.6 & 58.4 \\
 & AUPR~$\uparrow$ & 90.5 & 75.1 & 92.9 & 58.2 & {\cellcolor[rgb]{0.847,0.882,0.949}}98.9 & {\cellcolor[rgb]{0.545,0.647,0.839}}\textbf{99.7} & 98.1 &  & 59.5 & 92.8 & 32.8 & 65.7 &  & 64.5 & 28.2 \\
 & Sky-All~$\downarrow$ & 0.815 & 0.277 & 0.938 & 0.965 & 0.960 & 0.018 & 0.020 &  & {\cellcolor[rgb]{0.847,0.882,0.949}}0.005 & 0.283 & 0.992 & {\cellcolor[rgb]{0.847,0.882,0.949}}0.005 &  & {\cellcolor[rgb]{0.545,0.647,0.839}}\textbf{0.002} & 0.809 \\ 
\midrule
\multirow{3}{*}{4} & AUC~$\uparrow$ & 94.9 & 93.2 & 98.5 & 89.8 & 99.0 & {\cellcolor[rgb]{0.545,0.647,0.839}}\textbf{99.6} & {\cellcolor[rgb]{0.847,0.882,0.949}}99.5 &  & 87.2 & 96.1 & 58.8 & 91.8 &  & 71.3 & 58.4 \\
 & AUPR~$\uparrow$ & 87.2 & 71.1 & 92.4 & 59.8 & {\cellcolor[rgb]{0.847,0.882,0.949}}98.2 & {\cellcolor[rgb]{0.545,0.647,0.839}}\textbf{99.1} & 97.2 &  & 61.7 & 92.9 & 31.2 & 67.2 &  & 60.0 & 28.3 \\
 & Sky-All~$\downarrow$ & 0.868 & 0.284 & 0.940 & 0.945 & 0.959 & 0.023 & 0.022 &  & {\cellcolor[rgb]{0.847,0.882,0.949}}0.005 & 0.288 & 0.994 & {\cellcolor[rgb]{0.847,0.882,0.949}}0.005 &  & {\cellcolor[rgb]{0.545,0.647,0.839}}\textbf{0.002} & 0.819 \\ 
\midrule
\multirow{3}{*}{5} & AUC~$\uparrow$ & 92.5 & 90.3 & 97.6 & 89.6 & 98.2 & {\cellcolor[rgb]{0.847,0.882,0.949}}98.5 & {\cellcolor[rgb]{0.545,0.647,0.839}}\textbf{99.0} &  & 87.5 & 96.5 & 58.5 & 92.2 &  & 66.8 & 57.8 \\
 & AUPR~$\uparrow$ & 83.8 & 66.6 & 91.4 & 63.6 & {\cellcolor[rgb]{0.847,0.882,0.949}}96.8 & {\cellcolor[rgb]{0.545,0.647,0.839}}\textbf{97.1} & 96.1 &  & 64.6 & 93.7 & 32.8 & 70.4 &  & 53.8 & 28.2 \\
 & Sky-All~$\downarrow$ & 0.902 & 0.299 & 0.943 & 0.909 & 0.959 & 0.035 & 0.026 &  & {\cellcolor[rgb]{0.847,0.882,0.949}}0.005 & 0.295 & 0.996 & {\cellcolor[rgb]{0.847,0.882,0.949}}0.005 &  & {\cellcolor[rgb]{0.545,0.647,0.839}}\textbf{0.002} & 0.839 \\
\bottomrule
\end{tabular}
}
\caption{\textbf{Epistemic uncertainty estimation results encountering unseen pattern on Monocular depth estimation task.} The evaluation datasets used here are KITTI Seg-Depth (S$=$0) and KITTI Seg-Depth-C (S$>$0).}
\label{tab:epismde2_supp}
\end{table*}

\begin{figure}[t]
     \centering
         \includegraphics[width=0.42\textwidth]{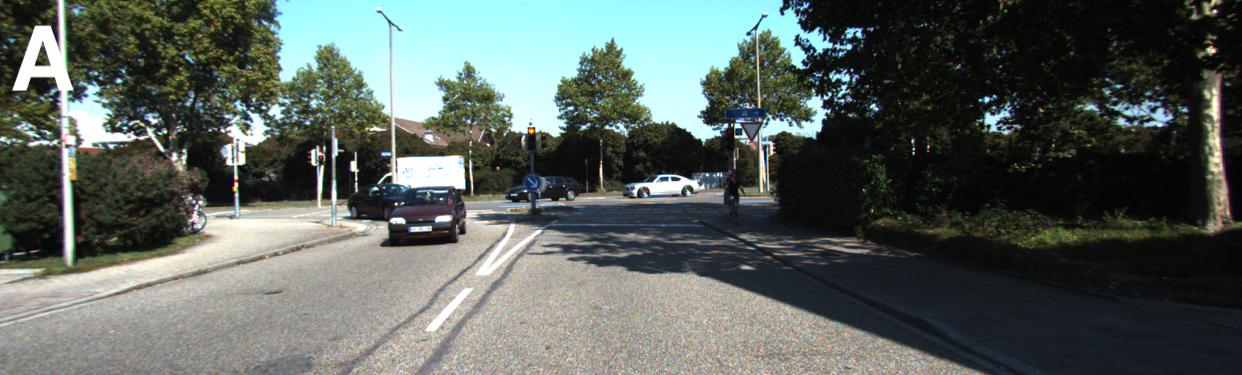}
         \includegraphics[width=0.42\textwidth]{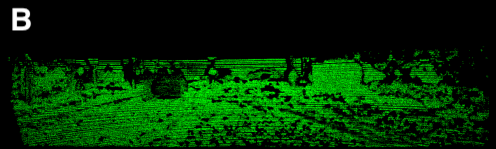}
         \includegraphics[width=0.42\textwidth]{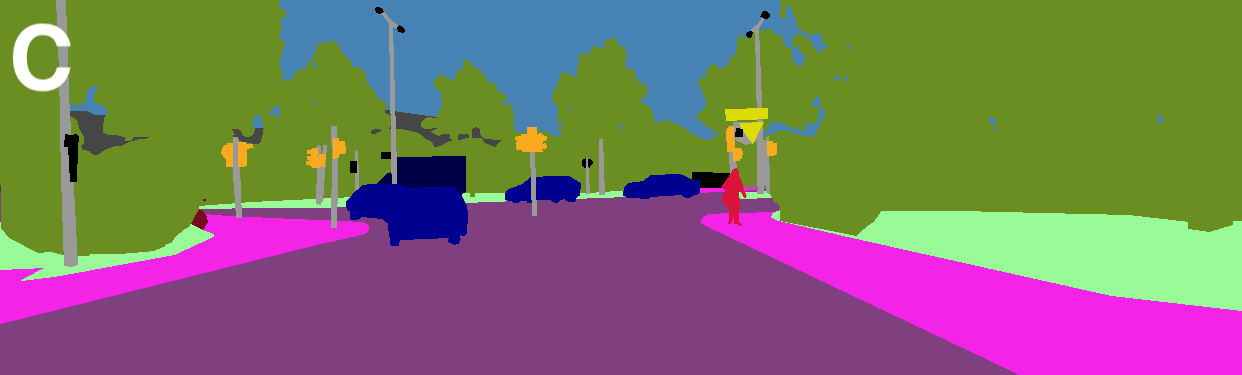}
         \includegraphics[width=0.42\textwidth]{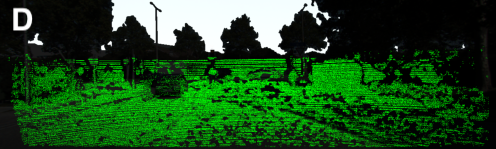}
        \vspace{1em}
        \caption{\textbf{Visualization on the OOD example detection in monocular depth estimation.} A: input image, B: ground truth depth map, green points represent pixels with depth ground truth; C: semantic ground truth map, blue areas represent pixels with sky pattern; D: ID-OOD map, green points represent the depth ground truth, i.e., the ID part. White parts represent the sky pattern, which is the OOD part.}
        \label{fig:sota_alea_eval_supp}
\end{figure}

\paragraph{A.5.6 Procedures and illustration for building sky pattern as the OOD examples}
In pixel-wise regression, when the main task model is applied to the same scenario as the one used in the training process, there will still be some OOD patterns in the image. For example, given an image shown Fig.~\ref{fig:sota_alea_eval_supp}-A, the corresponding depth ground truth map is provided in Fig.~\ref{fig:sota_alea_eval_supp}-B, with the annotated pixels colored in green. We can see that a big part of the pixels are not annotated by LIDAR because the orientation and the nature of LIDAR prevent it from covering all the ranges in the scene. We consider that the sky pattern is not annotated with depth value, but it can be labeled in the semantic map shown in Fig.~\ref{fig:sota_alea_eval_supp}-C. In this case, when we have a monocular depth estimation task, we can take the sky pattern as OOD examples and the pixels with depth annotations as the ID examples, as shown in Fig.~\ref{fig:sota_alea_eval_supp}-D, where the OOD patterns are colored in white and the ID patterns are colored in green. 

\section{Loss functions based on different distribution assumptions}
\label{supp:B}
\subsection{Loss functions}
We follow the notations in Sec 3 of the main paper and construct the loss functions based on different distribution assumptions. The following loss functions are used in the toy example, age estimation, and monocular depth estimation.
\paragraph{Laplace (Lap) distribution}
\begin{align}
&\mathcal{L}(\Theta_1) = \frac{1}{N}\sum_{i=1}^{N}\log(2 \sigma_{\Theta_1}({\mathbf{x}^{(i)}})) + \frac{| {y}^{(i)} - f_{\hat{\boldsymbol{\omega}}}(\mathbf{x}^{(i)})|}{\sigma_{\Theta_1}(\mathbf{x}^{(i)})}
\end{align}\noindent
we choose this distribution assumption for aleatoric uncertainty estimation in our generalized AuxUE solution. 

\paragraph{Generalized Gaussian (Ggau) distribution}
\begin{align}
\mathcal{L}(\Theta_1) &= \frac{1}{N}\sum_{i=1}^{N}\left( \frac{|{y}^{(i)} - f_{\hat{\boldsymbol{\omega}}}(\mathbf{x}^{(i)})|}{\hat{\alpha}^{(i)}} \right)^{\hat{\beta}^{(i)}} \nonumber\\
&- \log\frac{\hat{\beta}^{(i)}}{\hat{\alpha}^{(i)}}  + \log \Gamma (\frac{1}{\hat{\beta}^{(i)}})
\end{align}\noindent
with $\sigma_{\Theta_1}(\mathbf{x}^{(i)}) = (\hat{\alpha}^{(i)}, \hat{\beta}^{(i)})$, which means $\sigma_{\Theta_1}$ will output two other components (except for $\Tilde{y}^{(i)}$ defined in~\cite{upadhyay2022bayescap} which stands for $f_{\hat{\boldsymbol{\omega}}}(\mathbf{x}^{(i)})$ in our case) for Generalized Gaussian distribution. 

\paragraph{Normal Inverse Gamma (NIG) distribution}
\begin{align}
\mathcal{L}_1(\Theta_1) &= \frac{1}{N}\sum_{i=1}^{N}\frac{1}{2}\log(\frac{\pi}{\nu^{(i)}}) - \alpha^{(i)}\log(\Omega^{(i)}) \nonumber\\
&+ (\alpha^{(i)} + \frac{1}{2})\log((y^{(i)} - f_{\hat{\boldsymbol{\omega}}}(\mathbf{x}^{(i)}))^2\nu^{(i)} + \Omega^{(i)}) \nonumber\\
&+ \log(\frac{\Gamma(\alpha^{(i)})}{\Gamma(\alpha^{(i)} + \frac{1}{2})})
\end{align}\noindent

\begin{align}
&\mathcal{L}_2(\Theta_1) = \frac{1}{N}\sum_{i=1}^{N}|y^{(i)} - f_{\hat{\boldsymbol{\omega}}}(\mathbf{x}^{(i)})|\cdot(2\nu^{(i)} + \alpha^{(i)}) \nonumber\\
&\mathcal{L}(\Theta_1) = \mathcal{L}_1(\Theta_1) + \lambda_{\text{NIG}}\cdot\mathcal{L}_2(\Theta_1)
\end{align}\noindent
where $\Omega^{(i)} = 2\beta^{(i)}(1 + \nu^{(i)})$.
$\sigma_{\Theta_1}(\mathbf{x}^{(i)}) = (\hat{\alpha}^{(i)}, \hat{\beta}^{(i)}, \hat{\nu}^{(i)})$, which means $\sigma_{\Theta_1}$ will output three other components (except for $\gamma^{(i)}$ defined in~\cite{amini2020deep} which stands for $f_{\hat{\boldsymbol{\omega}}}(\mathbf{x}^{(i)})$ in our case) for Generalized Gaussian distribution. In our experiment, we set $\lambda_{\text{NIG}}=0.01$.

\subsection{Modifications in Super-resolution task}
\label{sec:modi_loss_supp}
In the BayesCap pipeline, the authors discover that, in their AuxUE, the reconstruction of the main task prediction can increase the uncertainty estimation performance. The reconstructed main task prediction is denoted as $\Tilde{\mathbf{y}}$.
We thus follow this idea in practice but only in the super-resolution experiment to have a fair comparison for different distribution assumptions. The loss function will have a slight difference and an additional identity mapping loss~\cite{upadhyay2022bayescap} than the proposed one in the main paper.
We list the modified loss functions as follows. The weight for the identity mapping loss $\lambda_{\text{Identity}}$ is set the same as in the original BayesCap.
\paragraph{Laplace (Lap) distribution}
\begin{align}
\mathcal{L}(\Theta_1) &= \frac{1}{N}\sum_{i=1}^{N}\log(2 \sigma_{\Theta_1}({\mathbf{x}^{(i)}})) + \frac{| {y}^{(i)} - \Tilde{y}^{(i)}|}{b^{(i)}} \nonumber\\
& + \lambda_{\text{Identity}} \cdot |f_{\hat{\boldsymbol{\omega}}}(\mathbf{x}^{(i)}) - \Tilde{y}^{(i)}|^2
\end{align}\noindent
where $\sigma_{\Theta_1}(\mathbf{x}^{(i)}) = (\Tilde{y}^{(i)}, b^{(i)})$.

\paragraph{Generalized Gaussian (Ggau) distribution}
\begin{align}
\mathcal{L}(\Theta_1) &= \frac{1}{N}\sum_{i=1}^{N}\left( \frac{|{y}^{(i)} - \Tilde{y}^{(i)}|}{\hat{\alpha}^{(i)}} \right)^{\hat{\beta}^{(i)}} - \log\frac{\hat{\beta}^{(i)}}{\hat{\alpha}^{(i)}} \nonumber\\
&  + \log \Gamma (\frac{1}{\hat{\beta}^{(i)}}) + \lambda_{\text{Identity}} \cdot |f_{\hat{\boldsymbol{\omega}}}(\mathbf{x}^{(i)}) - \Tilde{y}^{(i)}|^2
\end{align}\noindent
where $\sigma_{\Theta_1}(\mathbf{x}^{(i)}) = (\Tilde{y}^{(i)}, \hat{\alpha}^{(i)}, \hat{\beta}^{(i)})$.

\paragraph{Normal Inverse Gamma (NIG) distribution}
\begin{align}
\mathcal{L}_1(\Theta_1) &= \frac{1}{N}\sum_{i=1}^{N}\frac{1}{2}\log(\frac{\pi}{\nu^{(i)}}) - \alpha^{(i)}\log(\Omega^{(i)}) \nonumber\\
&+ (\alpha^{(i)} + \frac{1}{2})\log((y^{(i)} - \Tilde{y}^{(i)})^2\nu^{(i)} + \Omega^{(i)}) \nonumber\\
&+ \log(\frac{\Gamma(\alpha^{(i)})}{\Gamma(\alpha^{(i)} + \frac{1}{2})})
\end{align}\noindent

\begin{align}
\mathcal{L}_2(\Theta_1) = \frac{1}{N}\sum_{i=1}^{N}|y^{(i)} - \Tilde{y}^{(i)}|\cdot(2\nu^{(i)} + \alpha^{(i)})
\end{align}\noindent

\begin{align}
&\mathcal{L}(\Theta_1) = \mathcal{L}_1(\Theta_1) + \lambda_{\text{NIG}}\cdot\mathcal{L}_2(\Theta_1) \nonumber\\
&+ \lambda_{\text{Identity}}\cdot |f_{\hat{\boldsymbol{\omega}}}(\mathbf{x}^{(i)}) - \Tilde{y}^{(i)}|^2
\end{align}\noindent
where $\Omega^{(i)} = 2\beta^{(i)}(1 + \nu^{(i)})$, and $\sigma_{\Theta_1}(\mathbf{x}^{(i)}) = (\Tilde{y}^{(i)}, \hat{\alpha}^{(i)}, \hat{\beta}^{(i)}, \nu^{(i)})$. In our experiment, we set $\lambda_{\text{NIG}}=0.01$.

\section{Epistemic uncertainty estimation on AuxUE using prediction errors}\label{supp:C}
\subsection{Discretization for pixel-wise regression tasks}\label{supp:C1}
Given an image input $\mathbf{x}^{(i)}$, we can achieve an output map $\hat{\mathbf{y}}^{(i)}$. We consider the error map $\boldsymbol{\epsilon}^{(i)}$ contains $J^{(i)}$ valid pixels, and subscript $j$ as the indicator of the pixel. The values are sorted in ascending order, denoted by ${\boldsymbol{\epsilon}^{\prime}}^{(i)}$, with the same elements as $\boldsymbol{\epsilon}^{(i)}$. We divide ${\boldsymbol{\epsilon}^{\prime}}^{(i)}$ into $K$ subsets of equal size, represented by $\{\boldsymbol{\epsilon}^{(i)}_k\}_{k=1}^{K}$:
\begin{align}
\left\{\boldsymbol{\epsilon}_k^{(i)} \mid {\epsilon}^{(i)}_{\lfloor J^{(i)} * \frac{k-1}{K}\rceil} \leq {\epsilon}_j^{(i)}< {\epsilon}^{(i)}_{\lfloor J^{(i)} * \frac{k}{K}\rceil}\right\}^K_{k=1}
\end{align}\noindent
$\lfloor\cdot\rceil$ denotes the rounding operation. Each value in $\boldsymbol{\epsilon}_k^{(i)}$ is in the range of $\lfloor J^{(i)} *\frac{k-1}{K}\rceil$th and $\lfloor J^{(i)} *\frac{k}{K}\rceil$th value of the whole prediction error set $\boldsymbol{\epsilon}^{(i)}$.
Each error value ${\epsilon}_j^{(i)}$ is then replaced by the index of its corresponding subset $k \in [1, K]$ and transformed into a one-hot vector, denoted by $\bar{\boldsymbol{\epsilon}}_j^{(i)}$, as the final training target. Specifically, the one-hot vector is defined as:
\begin{align}
\small
\bar{\boldsymbol{\epsilon}}^{(i)}_{j} = [\bar{\epsilon}_{j,1}^{(i)} \ldots \bar{\epsilon}_{j,k}^{(i)} \ldots \bar{\epsilon}_{j,K}^{(i)}]^{\text{T}} \in \mathds{R}^K
\end{align}\noindent
where $\bar{\epsilon}^{(i)}_{j,k} = 1$ if $\epsilon_j^{(i)}$ belongs to the $k$th subset, and 0 otherwise.

\subsection{Demo code for discretization operation}\label{supp:C2}
For the sake of clarification, we provide the demonstration code in DemoCode~\ref{discretization_dido_supp} directly in Python and PyTorch. The function \texttt{discretization\_imagelevel} and \texttt{discretization\_pixelwise} represent the discretization for image-level and pixel-wise tasks, respectively. Note that some modifications might be needed when deploying them in practice.

\subsection{Modeling epistemic uncertainty using prediction errors}\label{supp:C3}
In a Bayesian framework, given an input $\mathbf{x}$, the predictive uncertainty of a DNN is modeled by $P(y|\mathbf{x}, \mathcal{D})$. We can have the following assumptions and simplifications. Since we have a trained main task DNN, and as proposed in~\cite{malinin2018predictive}, we assume \textit{a point-estimate of $\boldsymbol{\omega}$}, and in auxiliary uncertainty estimation case, we have the trained and fixed main task model with parameters $\hat{\boldsymbol{\omega}}$, then we have:
\begin{align}
P(\boldsymbol{\omega}|\mathcal{D}) = \delta(\boldsymbol{\omega} - \hat{\boldsymbol{\omega}}) \rightarrow
P(y|\mathbf{x}, \mathcal{D}) \approx P(y|\mathbf{x}, {\hat{\boldsymbol{\omega}}})
\label{eq:transform1_dido_supp}
\end{align}\noindent
with $\delta$ being the Dirac function.

We then follow the Gaussian assumption, i.e., the prediction is drawn from $\mathcal{N}(y| {\mu}, {\sigma}^2)$ and according to the modeling in evidential regression~\cite{amini2020deep}, we denote $\boldsymbol{\alpha}$ as the parameters of prior distributions of $({\mu}, {\sigma}^{2})$. Following the same work, we first have:
\begin{align}
P(\mu, \sigma^2| \mathbf{x}, \boldsymbol{\alpha}, \hat{\boldsymbol{\omega}}) = P(\mu|\sigma^2, \mathbf{x}, \boldsymbol{\alpha}, \hat{\boldsymbol{\omega}})P(\sigma^2| \mathbf{x}, \boldsymbol{\alpha},\hat{\boldsymbol{\omega}})\label{eq:amini}
\end{align}\noindent
According to Eq.~\ref{eq:transform1_dido_supp}, we regard the $\mu$ depends only on $\mathbf{x}$ and the main task model $\hat{\boldsymbol{\omega}}$:
\begin{align}
P(\mu, \sigma^2| \mathbf{x}, \boldsymbol{\alpha}, \hat{\boldsymbol{\omega}}) &= P(\mu| \mathbf{x}, \hat{\boldsymbol{\omega}})P(\sigma^2| \mathbf{x},\boldsymbol{\alpha},\hat{\boldsymbol{\omega}})\nonumber\\
&= \delta(\mu - f_{\hat{\boldsymbol{\omega}}}(\mathbf{x}))P(\sigma^2| \mathbf{x},\boldsymbol{\alpha},\hat{\boldsymbol{\omega}})\label{eq:my}
\end{align}\noindent
We introduce $\boldsymbol{\alpha}$ and re-write $P(y|\mathbf{x}, \hat{\boldsymbol{\omega}})$ in Eq.~\ref{eq:transform1_dido_supp} as:
\begin{small}
\begin{align}
    P(y|\mathbf{x}, \boldsymbol{\alpha}, \hat{\boldsymbol{\omega}}) &= \iint P(y|\mu, \sigma^{2}) P(\mu, \sigma^{2} | \mathbf{x}, \boldsymbol{\alpha}, \hat{\boldsymbol{\omega}}) d{\mu}d{\sigma^{2}}
    \nonumber\\
    &\labelrel={eq:amini2} \iint P(y|\mu, \sigma^{2}) P(\mu | \sigma^{2}, \mathbf{x}, \boldsymbol{\alpha}, \hat{\boldsymbol{\omega}}) P(\sigma^{2} | \mathbf{x}, \boldsymbol{\alpha}, \hat{\boldsymbol{\omega}}) d{\mu}d{\sigma^{2}}    \nonumber\\
    &= \iint P(y, \mu| \sigma^{2}, \mathbf{x},\boldsymbol{\alpha}, \hat{\boldsymbol{\omega}}) P(\sigma^{2} | \mathbf{x},\boldsymbol{\alpha}, \hat{\boldsymbol{\omega}}) d{\mu}d{\sigma^{2}}    \nonumber\\
    &\labelrel={eq:mysimp2} \int \delta(\mu - f_{\hat{\omega}}(\mathbf{x})) d{\mu} \int P(y|\sigma^{2}, \mathbf{x}, \boldsymbol{\alpha}, \hat{\boldsymbol{\omega}}) P(\sigma^{2} | \mathbf{x}, \boldsymbol{\alpha}, \hat{\boldsymbol{\omega}}) d{\sigma^{2}} \nonumber\\
    &= \int P(y|\mathbf{x}, \sigma^{2}) P(\sigma^{2} | \mathbf{x}, \boldsymbol{\alpha}, \hat{\boldsymbol{\omega}}) d{\sigma^{2}}
    \label{eq:predictive_u_dido_supp}
\end{align}
\end{small}\noindent
where the equality~\eqref{eq:amini2} and \eqref{eq:mysimp2} in Eq.~\ref{eq:predictive_u_dido_supp} are given by Eq.~\ref{eq:amini} and Eq.~\ref{eq:my}, respectively.

In summary, we have 
\begin{align}
    P(y| \mathbf{x}, \mathcal{D}) &= \iint P(y | \mathbf{x}, \sigma^2) P(\sigma^2 | \boldsymbol{\omega}) P(\boldsymbol{\omega} | \mathcal{D}) d\sigma^2 d\boldsymbol{\omega} \nonumber\\
    &= \int P(y | \mathbf{x}, \sigma^2) P(\sigma^2 | \mathcal{D}) d\sigma^2 \nonumber\\
    &\labelrel\approx{eq:mysimp3} \int P(y|\mathbf{x}, \sigma^{2}) P(\sigma^{2} | \mathbf{x}, \boldsymbol{\alpha}, \hat{\boldsymbol{\omega}}) d{\sigma^{2}}
    \label{eq:final_dido}
\end{align}\noindent
where the approximation~\eqref{eq:mysimp3} in Eq.~\ref{eq:final_dido} is given by Eq.~\ref{eq:predictive_u_dido_supp}.

\begin{figure}[t]
\centering
\includegraphics[width=0.46\textwidth]{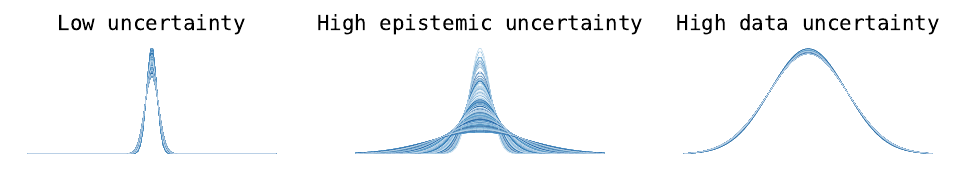}
\caption{\textbf{Visualizations on desired behaviors of regression results in auxiliary uncertainty estimation scenario.} {Different types of uncertainty result in different distributions of the variance under Gaussian assumption.}}
\label{fig:vis_simple_dido_supp}
\end{figure}

In this case, we first consider $\epsilon$ to be drawn from a continuous distribution parameterized by $\sigma^2$. Furthermore, we argue that $P(\sigma^2 | \mathcal{D})$ describes the epistemic uncertainty when we have a trained and fixed main task model and the variational approach can be applied~\cite{joo2020being,malinin2018predictive}: $P( \sigma^2 | \mathbf{x}, \boldsymbol{\alpha}, \hat{\boldsymbol{\omega}}) \approx P(\sigma^2 | {\mathcal{D}})$. 
It shows the special case of the approximation for the posterior over $y$, where the mean is fixed and only variances differ. Similar to~\cite{malinin2020regression}, we can illustrate this by using the ensembles of the regression results as in Figure~\ref{fig:vis_simple_dido_supp}. 
The discrepancy in variances determines the epistemic uncertainty of the final prediction.

\begin{figure*}
\lstset{style=mystyle}
\begin{lstlisting}[language=Python, caption=\textbf{Discretization code in Python}. Both image-level and pixel-wise cases are provided. Note that the exact code in practice might need some modifications based on this., label=discretization_dido_supp]
import torch
import torch.nn.functional as F

def discretization_imagelevel(data_loader, x, y, f_omega, K):
    """Discretization operation on image-level tasks.
    Args:
        data_loader: training or validation set data loader
        x: input image
        y: ground truth target
        f_omega: trained main DNN
        K: number of classes
    Returns:
        epsilon_bar: one-hot prediction error
    """
    epsilon = []
    for x_item, y_item in data_loader:
        epsilon.append(abs(y_item - f_omega(x_item)))
    epsilon = torch.cat(epsilon, dim=0)
    quantiles = torch.quantile(epsilon, torch.tensor(range(K+1)/K))
    classes = torch.tensor(range(K))
    for i, (q1, q2, c) in enumerate(zip(quantiles[:-1], quantiles[1:], classes)):
        if i == 0: 
            mask_q1 = epsilon >= q1
        else: 
            mask_q1 = epsilon > q1
        mask_q2 = epsilon <= q2
        mask_q = torch.logical_and(mask_q1, mask_q2)
        temp[mask_q] = c
    epsilon_bar = F.one_hot(temp, K)
    return epsilon_bar
    
def discretization_pixelwise(x, y, f_omega, K):
    """Discretization operation on pixel-wise tasks.
    Args:
        x: input image
        y: ground truth map
        f_omega: trained main DNN
        K: number of classes
    Returns:
        epsilon_bar: one-hot prediction error map
    """
    epsilon = abs(y - f_omega(x))
    temp = torch.zeros_like(epsilon)
    quantiles = torch.quantile(epsilon, torch.tensor(range(K+1)/K))
    classes = torch.tensor(range(K))
    for i, (q1, q2, c) in enumerate(zip(quantiles[:-1], quantiles[1:], classes)):
        if i == 0: 
            mask_q1 = epsilon >= q1
        else: 
            mask_q1 = epsilon > q1
        mask_q2 = epsilon <= q2
        mask_q = torch.logical_and(mask_q1, mask_q2)
        temp[mask_q] = c
    epsilon_bar = F.one_hot(temp, K)
    return epsilon_bar
\end{lstlisting}
\end{figure*}

\section{Ablation study}\label{supp:D}
The ablation studies are based on the monocular depth estimation task.
\subsection{Hyperparameters}
There are two main hyperparameters in our proposed DIDO: the number of the sets $K$ in discretization and $\lambda$ for the regularization term in the loss function (Eq. 4. in the main paper). In this section, we analyze the effect of these two hyperparameters.

The evaluations are based on epistemic uncertainty estimation on unseen patterns and dataset change detection. 

\begin{figure*}[t]
\centering
    \begin{subfigure}{\textwidth}
        \centering
        \includegraphics[width=0.32\linewidth]{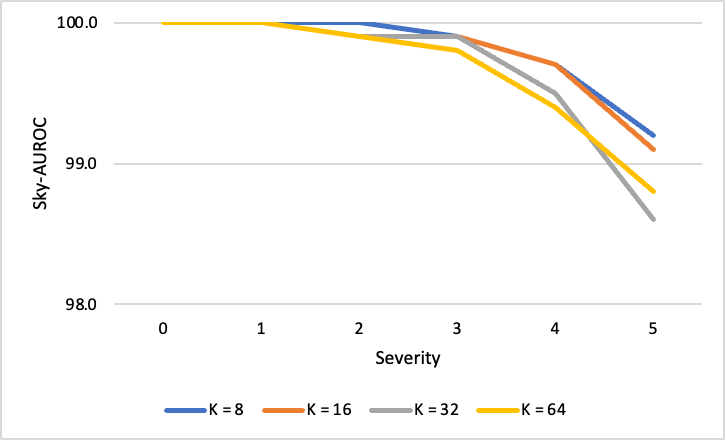}
        \includegraphics[width=0.32\linewidth]{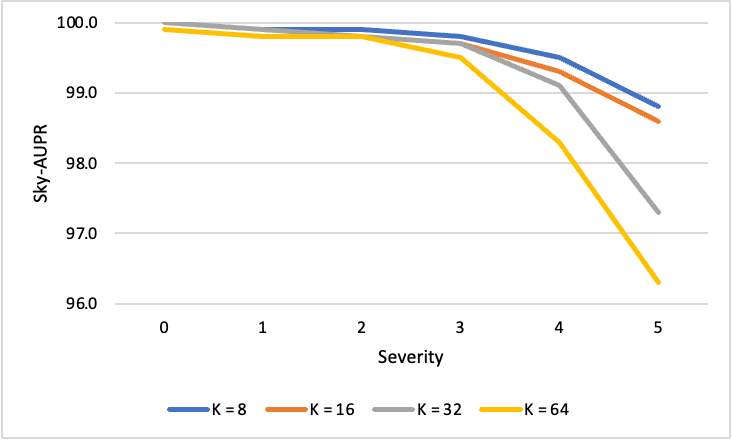}
        \includegraphics[width=0.32\linewidth]{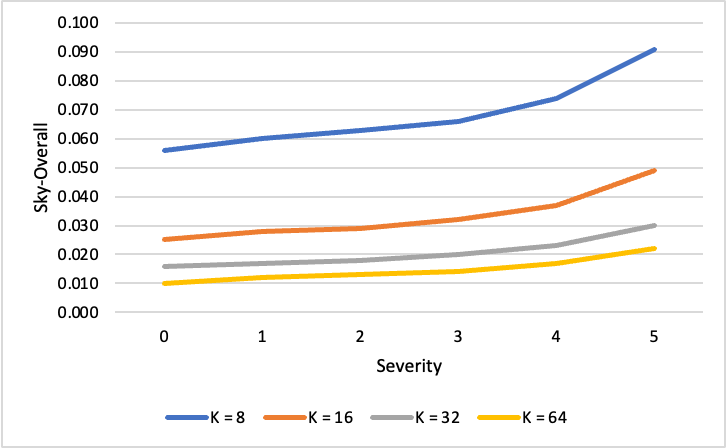}
        \caption{\textbf{Ablation study on $K$ for DIDO on unseen patterns detection in KITTI dataset.} The results are given by DIDO-based AuxUE with different numbers of classes ($K$) in discretization. }
        \label{tab:abla_1_a}
    \end{subfigure}
    \begin{subfigure}{\textwidth}
        \centering
        \includegraphics[width=0.32\linewidth]{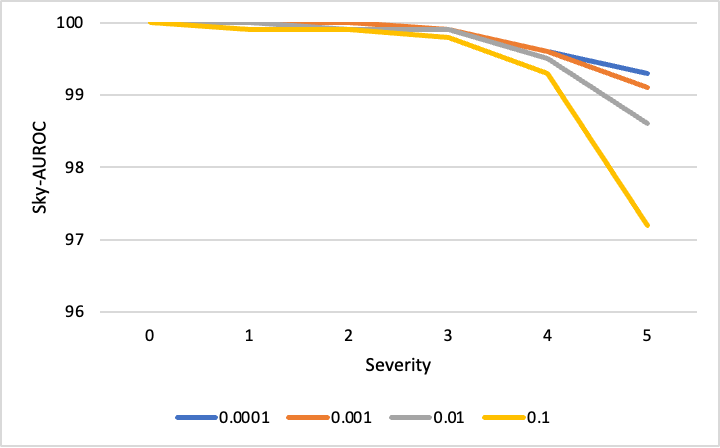}
        \includegraphics[width=0.32\linewidth]{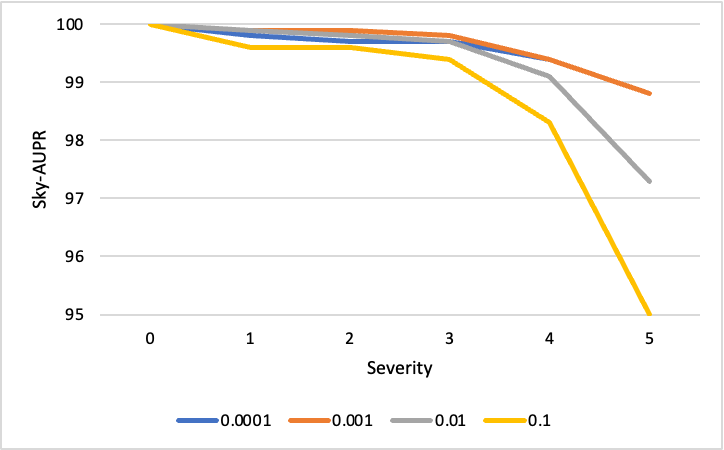}
        \includegraphics[width=0.32\linewidth]{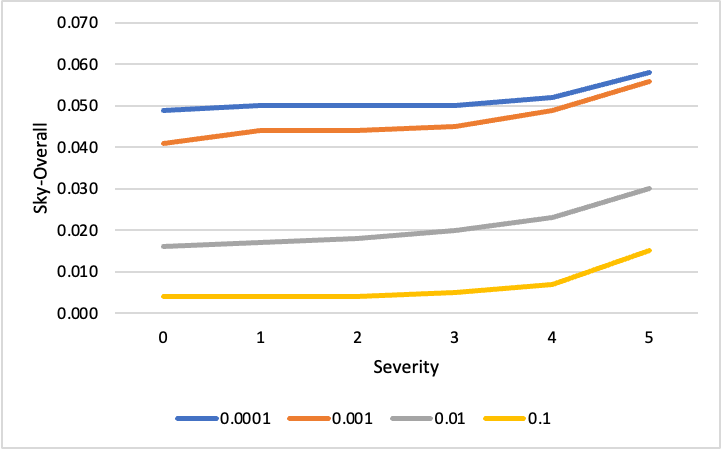}
        \caption{\textbf{Ablation study on $\lambda$ for DIDO on unseen patterns detection in KITTI dataset.} The results are given by DIDO-based AuxUE with ($K = 32$) trained by using different $\lambda$ for the regularization term in loss $L(\Theta_2)$.}
        \label{tab:abla_1_b}
    \end{subfigure}
    \begin{subfigure}{\textwidth}
        \centering
        \includegraphics[width=0.5\linewidth]{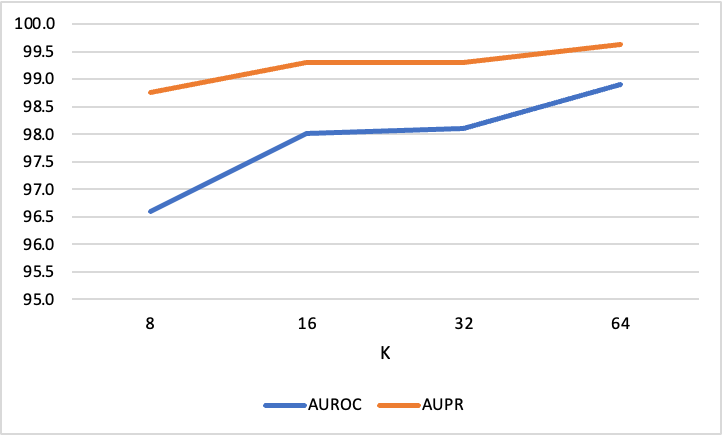}
        \caption{\textbf{Ablation study on $K$ for DIDO on dataset change detection in monocular depth estimation.} The evaluation is made by taking KITTI outdoor dataset as the In-Distribution data and NYU indoor dataset as the Out-of-Distribution data.}
        \label{tab:abla_1_c}
    \end{subfigure}
    \begin{subfigure}{\textwidth}
        \centering
        \includegraphics[width=0.5\linewidth]{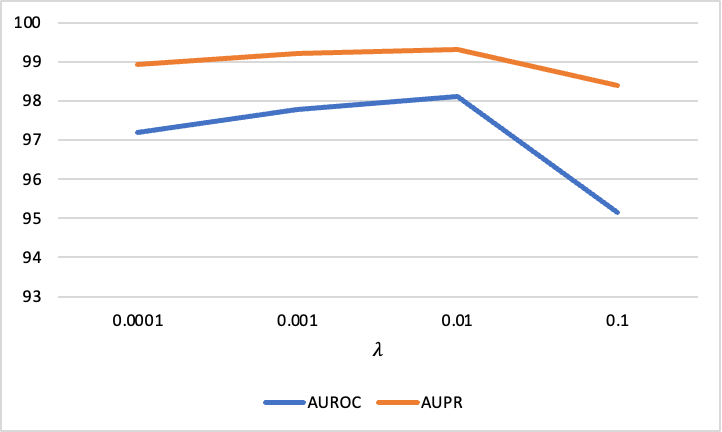}
        \caption{\textbf{Ablation study on $\lambda$ for DIDO on dataset change detection in monocular depth estimation.} The evaluation is made by taking KITTI outdoor dataset as the In-Distribution data and NYU indoor dataset as the Out-of-Distribution data.}
        \label{tab:abla_1_d}
    \end{subfigure}
\caption{\textbf{Ablation study on hyperparameters for DIDO on monocular depth estimation.}}
\label{tab:abla_1}
\end{figure*}

The effect of $K$ is shown in Fig.~\ref{tab:abla_1_a} and Fig.~\ref{tab:abla_1_c}. We test $K=\{8,16,32,64\}$ and $\lambda$ is fixed to $0.01$. The AUC and AUPR performance decrease when we have bigger $K$, while on the Sky-All metric, bigger $K$ provides better results. When the evaluation dataset is changed from KITTI to NYU~\cite{Silberman:ECCV12}, bigger $K$ can also provide better AUC and AUPR in identifying the change. We choose $K=32$ to have a balanced performance. 

The effect of $\lambda$ is shown in Fig.~\ref{tab:abla_1_b} and Fig.~\ref{tab:abla_1_d}. We test $\lambda=\{1e-4,1e-3,1e-2,1e-1\}$ and $K$ is fixed to 32. We can see the AUC and AUPR performance decrease when we use bigger $\lambda$ during training, while on the Sky-All metric, it performs better when using bigger $\lambda$ during training. For the dataset change experiment, we can see when $\lambda=0.01$, the DIDO can provide the best AUC and AUPR. We choose $\lambda=0.01$ in the end for our model.

\subsection{Necessity of using AuxUE}
In this experiment, we apply DIDO on the main task BTS~\cite{lee2019big} model to see the impact on the main task performance and the uncertainty estimation performance. The comparison will only be conducted with the other modified main task models for fairness.
In particular, we first adjust the BTS model to Single Predictive Uncertainty~\cite{nix1994uncertainty,kendall2017uncertainties} variant (BTS-SinglePU), i.e., we first add an aleatoric uncertainty estimation head parallel to and identical to the depth estimation head on top of the original model. Then we add the same head for DIDO applied on $\sigma_{\Theta_2}$. Thus there are three prediction heads on the modified BTS model corresponding to depth prediction, aleatoric uncertainty estimation and epistemic uncertainty estimation. We denote this variant as BTS-DIDO. We will compare BTS-DIDO with the original BTS model (Org) and the BTS-SinglePU model to check the impact on the main task. We also compare the BTS-DIDO with AuxUE + original BTS, BTS-DEns. and BTS-SinglePU models to check the uncertainty estimation performance. 

To train the BTS-SinglePU models, we follow the original BTS settings for hyperparameters in training. We change the loss function to Gaussian NLL loss. For BTS-DIDO, we use the same hyperparameters as we used in AuxUEs for DIDO modeling, i.e., $K=32$ and $\lambda=0.01$. For the other hyperparameters, such as the batch size and learning rate, we follow the original BTS settings. 

During training, we found that combining DIDO directly with the BTS will make the training unstable: the loss will be exploded after around fifteen epochs. As shown in Tab.~\ref{tab:abla_btsdido_1}, for the main task performance, the original BTS can outperform the others even for BTS-DEns. We argue that there are two reasons that might result in the performance reduction: BTS-DEns. component models (BTS-SinglePU models) are adjusted for the uncertainty output; SiLog loss~\cite{eigen2014depth}, which is specifically applied to the MDE task, is replaced by the Gaussian negative log-likelihood loss. However, when the noise severity increases ($S>3$), BTS-DIDO and BTS-DEns. can perform better than the others. In particular, BTS-DIDO shows a more robust performance given the inputs with heavy perturbations. 

For the uncertainty estimation performance, as shown in Tab.~\ref{tab:abla_btsdido_2}, BTS-DIDO and AuxUE achieve similar performance on AUC and AUPR metrics and on dataset change detection. While on Sky-All, AuxUE works slightly better than BTS-DIDO. For aleatoric uncertainty, since the main task performances for different models are different, the comparison can only be a reference. With a sacrifice on the main task performance, BTS-DIDO has the potential to achieve good uncertainty estimation performance.

We argue that it is necessary to use AuxUE to keep the main task performance when the input is relatively clean. Meanwhile, the good performance on BTS-DIDO under high severity perturbations makes it meaningful to work on stabilizing the training for DIDO-based models in the future.

\subsection{Effectiveness of Dirichlet modeling}
\begin{figure*}[t]
\centering
        \includegraphics[width=0.32\linewidth]{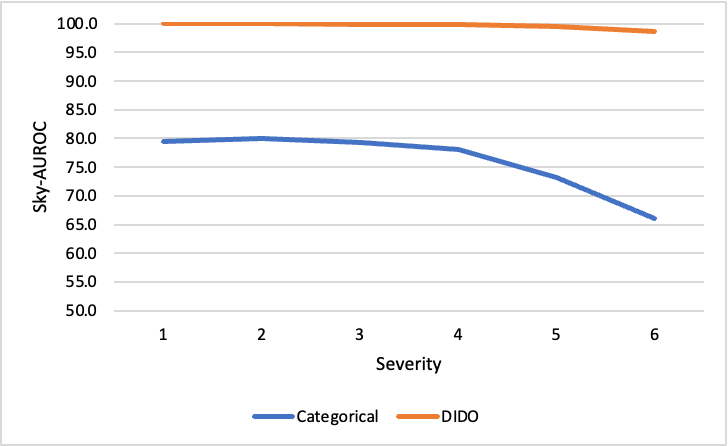}
        \includegraphics[width=0.32\linewidth]{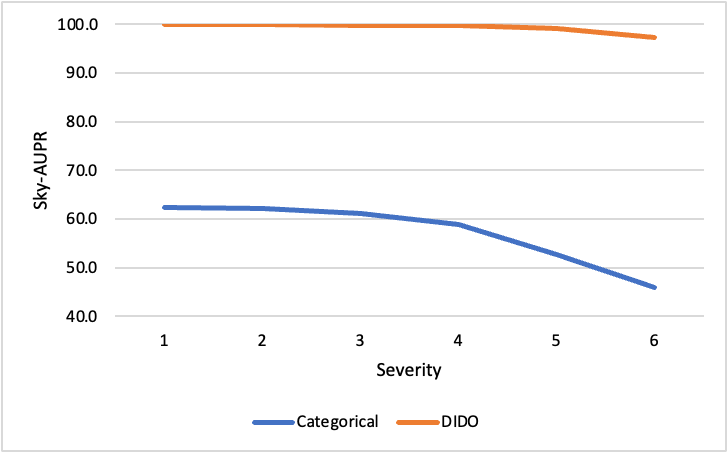}
        \includegraphics[width=0.32\linewidth]{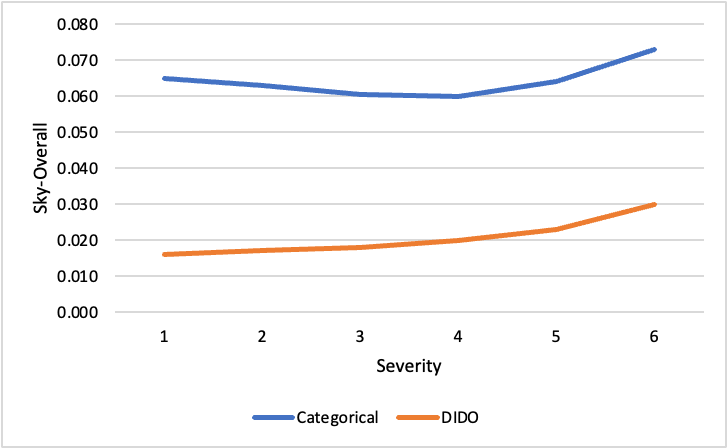}
\caption{\textbf{Ablation study on the effectiveness of Dirichlet modeling for DIDO on monocular depth estimation.} $K=32$ for both Categorical and Dirichlet modeling cases.}
\label{tab:abla_cate}
\end{figure*}

We show the effectiveness of the Dirichlet modeling instead of using the standard categorical modeling based on discretized prediction errors.
For categorical modeling, we also choose $K=32$ classes for discretization. We change the activation function on the top of $\sigma_{\Theta_2}$ from the ReLU function to the Softmax function, then apply classical cross-entropy on the Softmax outputs. For measuring uncertainty, we use the Shannon-Entropy~\cite{shannon2001mathematical} on the Softmax outputs. As shown in Fig.~\ref{tab:abla_cate}, the Dirichlet modeling outperforms the Categorical modeling on all three metrics w.r.t. the OOD pattern detection. On the dataset change experiment, Categorial modeling provides 90.37 for AUC and 96.83 for AUPR, which underperforms the results given by Dirichlet modeling. This study shows the effectiveness of DIDO and the use of evidential learning in the AuxUE.

\section{More visualizations}\label{supp:E}
Fig.~\ref{tab:vis_mde_supp} shows more visualizations on monocular depth estimation. For aleatoric uncertainty estimation maps, since some of the values on the unseen part (mostly the upper part of the map) are extremely high ($>$1e4), we \textbf{clip the values} to the maximum predicted value on the pixels with ground truth for better illustration. As uncertainty estimates show, our proposed DIDO can highlight the patterns rarely appearing throughout the whole dataset, e.g., the windshield of the car, the underside of the car, the barbed wire fence, and the upper part of the image like the sky. 
However, only the sky part is a pattern that must have no ground truth depth values and have semantic segmentation annotations. This is the reason we chose only the sky as the OOD pattern.
Note that DIDO won't always highlight the areas without ground truth. For instance, we can see DIDO does not always assign higher uncertainty on the parts with no ground truth for the body of the car or the road, since some of these patterns might have ground truth on the other images in the training set or share similar patterns which have ground truth values.

\begin{table*}[t]
\centering
\scalebox{0.8}{
\begin{tabular}{clcccccccc} 
\toprule
\multicolumn{1}{l}{} &  & \multicolumn{8}{c}{Main task performance} \\ 
\cmidrule{3-10}
S & Methods & absrel $\downarrow$ & log10 $\downarrow$ & rms $\downarrow$ & sqrel $\downarrow$ & logrms $\downarrow$ & d1 $\uparrow$ & d2 $\uparrow$ & d3 $\uparrow$ \\ 
\toprule
\multirow{4}{*}{0} & Org + AuxUE & \first \textbf{0.056} & \first \textbf{0.025} & \first \textbf{2.430} & \first \textbf{0.201} & \first \textbf{0.089} & \first \textbf{0.963} & \second {0.994} & \first \textbf{0.999} \\
 & BTS-SinglePU & 0.065 & 0.029 & 2.606 & 0.234 & 0.100 & 0.952 & 0.993 & \second {0.998} \\
 & BTS-DEns. & \second {0.060} & \second {0.026} & \second {2.435} & \second {0.202} & \second {0.092} & \second {0.961} & \first \textbf{0.995} & \first \textbf{0.999} \\
 & BTS-DIDO & 0.061 & 0.027 & 2.574 & 0.236 & 0.098 & 0.954 & 0.992 & 0.998 \\ 
\midrule
\multirow{4}{*}{1} & Org + AuxUE & \first \textbf{0.077} & \first \textbf{0.036} & \first \textbf{3.185} & \first \textbf{0.370} & \first \textbf{0.129} & \first \textbf{0.919} & \first \textbf{0.977} & \first \textbf{0.992} \\
 & BTS-SinglePU & 0.094 & 0.043 & 3.581 & 0.476 & 0.149 & 0.890 & 0.969 & 0.989 \\
 & BTS-DEns. & \second {0.087} & \second {0.040} & \second {3.415} & \second {0.422} & \second {0.138} & \second {0.902} & \second {0.974} & \first \textbf{0.992} \\
 & BTS-DIDO & 0.088 & 0.040 & 3.453 & 0.456 & 0.143 & 0.898 & 0.972 & 0.991 \\ 
\midrule
\multirow{4}{*}{2} & Org + AuxUE & \first \textbf{0.096} & \first \textbf{0.047} & \first \textbf{3.861} & \first \textbf{0.571} & \first \textbf{0.168} & \first \textbf{0.876} & \first \textbf{0.954} & \second {0.979} \\
 & BTS-SinglePU & 0.116 & 0.057 & 4.359 & 0.735 & 0.192 & 0.835 & 0.939 & 0.973 \\
 & BTS-DEns. & 0.109 & 0.053 & 4.189 & \second {0.661} & \second {0.178} & 0.848 & 0.947 & \second {0.979} \\
 & BTS-DIDO & \second {0.108} & \second {0.051} & \second {4.169} & 0.670 & \second {0.178} & \second {0.851} & \second {0.948} & \first \textbf{0.980} \\ 
\midrule
\multirow{4}{*}{3} & Org + AuxUE & \first \textbf{0.130} & \second {0.069} & \first \textbf{4.905} & \first \textbf{0.985} & 0.237 & \first \textbf{0.805} & \second {0.908} & 0.949 \\
 & BTS-SinglePU & 0.149 & 0.078 & 5.357 & 1.140 & 0.253 & 0.760 & 0.890 & 0.944 \\
 & BTS-DEns. & 0.140 & 0.073 & 5.184 & 1.031 & 0.234 & 0.772 & 0.904 & 0.955 \\
 & BTS-DIDO & \second {0.134} & \textbf{\first \textbf{0.067}} & \second {5.134} & \second {1.003} & \first \textbf{0.228} & \second {0.789} & \first \textbf{0.912} & \first \textbf{0.961} \\ 
\midrule
\multirow{4}{*}{4} & Org + AuxUE & 0.195 & 0.117 & 6.591 & 1.888 & 0.370 & \second {0.680} & 0.808 & 0.874 \\
 & BTS-SinglePU & 0.195 & 0.110 & 6.649 & 1.786 & 0.341 & 0.662 & 0.816 & 0.894 \\
 & BTS-DEns. & \second {0.186} & \second {0.103} & \second {6.485} & \second {1.649} & \second {0.317} & 0.667 & \second {0.833} & \second {0.911} \\
 & BTS-DIDO & \first \textbf{0.170} & \first \textbf{0.089} & \first \textbf{6.292} & \first \textbf{1.485} & \first \textbf{0.293} & \first \textbf{0.711} & \first \textbf{0.862} & \first \textbf{0.930} \\ 
\midrule
\multirow{4}{*}{5} & Org + AuxUE & 0.265 & 0.172 & 8.259 & 2.932 & 0.508 & 0.555 & 0.696 & 0.783 \\
 & BTS-SinglePU & 0.231 & 0.135 & 7.731 & 2.328 & 0.410 & 0.585 & 0.757 & 0.853 \\
 & BTS-DEns. & \second {0.222} & \second {0.127} & \second {7.584} & \second {2.190} & \second {0.386} & \second {0.587} & \second {0.772} & \second {0.871} \\
 & BTS-DIDO & \first \textbf{0.211} & \first \textbf{0.116} & \first \textbf{7.484} & \first \textbf{2.065} & \first \textbf{0.367} & \first \textbf{0.621} & \first \textbf{0.799} & \first \textbf{0.890} \\
\toprule
\end{tabular}
}
\caption{\textbf{Ablation study on the necessity of using AuxUE.} Main task performance comparison on KITTI and KITTI-C.}
\label{tab:abla_btsdido_1}
\end{table*}

\begin{table*}[t]
\centering
\scalebox{0.73}{
\begin{tabular}{clccccccc|cc}
\multicolumn{1}{l}{} &  & \multicolumn{4}{c}{Aleatoric uncertainty estimation} & \multicolumn{3}{c}{Epsitemic uncertainty: Unseen pattern} & \multicolumn{2}{c}{Dataset change} \\ 
\cmidrule{3-11}
S & Methods & AUSE-REL $\downarrow$ & AUSE-RMSE $\downarrow$ & AURG-REL $\uparrow$ & AURG-RMSE $\uparrow$ & AUC $\uparrow$ & AUPR $\uparrow$ & \multicolumn{1}{c}{Sky-All $\downarrow$} & AUC $\uparrow$ & AUPR $\uparrow$ \\ 
\toprule
\multirow{4}{*}{0} & Org + AuxUE & \first \textbf{0.013} & \second {0.203} & 0.023 & 1.870 & \first \textbf{100.0} & \first \textbf{100.0} & \multicolumn{1}{c}{\second {0.015}} & \second {98.1} & \second {99.3} \\
 & BTS-SinglePU & 0.016 & 0.222 & \second {0.026} & \second {1.978} & 89.0 & 62.0 & \multicolumn{1}{c}{\first \textbf{0.005}} & 64.2 & 78.3 \\
 & DEns. & \second {0.014} & \first \textbf{0.195} & 0.024 & 1.866 & \second {93.5} & \second {70.0} & \multicolumn{1}{c}{\first \textbf{0.005}} & 62.1 & 76.7 \\
 & BTS-DIDO & \first \textbf{0.013} & 0.207 & \first \textbf{0.028} & \first \textbf{1.990} & \first \textbf{100.0} & \first \textbf{100.0} & \multicolumn{1}{c}{0.017} & \first \textbf{98.5} & \first \textbf{99.5} \\ 
\midrule
\multirow{4}{*}{1} & Org + AuxUE & \second {0.019} & 0.336 & 0.031 & 2.361 & \first \textbf{100.0} & \first \textbf{99.9} & \second {0.016} & \multicolumn{2}{c}{\multirow{20}{*}{-}} \\
 & BTS-SinglePU & 0.021 & 0.330 & \second {0.038} & \first \textbf{2.657} & 86.9 & 59.1 & \first \textbf{0.005} & \multicolumn{2}{c}{} \\
 & BTS-DEns. & \second {0.019} & \first \textbf{0.285} & 0.036 & 2.573 & 92.8 & \second {68.0} & \first \textbf{0.005} & \multicolumn{2}{c}{} \\
 & BTS-DIDO & \first \textbf{0.017} & \second {0.308} & \first \textbf{0.041} & \second {2.608} & \first \textbf{100.0} & \first \textbf{99.9} & 0.027 & \multicolumn{2}{c}{} \\ 
\cmidrule{1-9}
\multirow{4}{*}{2} & Org + AuxUE & 0.023 & 0.468 & 0.038 & 2.774 & \second {99.9} & \second {99.8} & 0.017 & \multicolumn{2}{c}{} \\
 & BTS-SinglePU & 0.026 & 0.443 & \second {0.046} & \first \textbf{3.150} & 86.6 & 58.9 & \first \textbf{0.005} & \multicolumn{2}{c}{} \\
 & BTS-DEns. & \second {0.022} & \first \textbf{0.387} & 0.044 & 3.078 & 92.3 & 67.0 & \first \textbf{0.005} & \multicolumn{2}{c}{} \\
 & BTS-DIDO & \first \textbf{0.021} & \second {0.396} & \first \textbf{0.050} & \second {3.093} & \first \textbf{100.0} & \first \textbf{99.9} & 0.033 & \multicolumn{2}{c}{} \\ 
\cmidrule{1-9}
\multirow{4}{*}{3} & Org + AuxUE & 0.031 & 0.730 & 0.049 & 3.308 & \first \textbf{99.9} & \second {99.7} & \second {0.018} & \multicolumn{2}{c}{} \\
 & BTS-SinglePU & 0.031 & 0.619 & \second {0.055} & \second {3.719} & 86.6 & 59.5 & \first \textbf{0.005} & \multicolumn{2}{c}{} \\
 & BTS-DEns. & \second {0.027} & \second {0.526} & 0.054 & 3.685 & \second {91.6} & 65.7 & \first \textbf{0.005} & \multicolumn{2}{c}{} \\
 & BTS-DIDO & \first \textbf{0.023} & \first \textbf{0.500} & \first \textbf{0.062} & \first \textbf{3.749} & \first \textbf{99.9} & \first \textbf{99.8} & 0.036 & \multicolumn{2}{c}{} \\ 
\cmidrule{1-9}
\multirow{4}{*}{4} & Org + AuxUE & 0.049 & 1.268 & 0.059 & 3.929 & \second {99.6} & \second {99.1} & \second {0.023} & \multicolumn{2}{c}{} \\
 & BTS-SinglePU & 0.038 & 0.905 & \second {0.067} & 4.345 & 87.2 & 61.7 & \first \textbf{0.005} & \multicolumn{2}{c}{} \\
 & BTS-DEns. & \second {0.032} & \second {0.734} & \second {0.067} & \second {4.401} & 91.8 & 67.2 & \first \textbf{0.005} & \multicolumn{2}{c}{} \\
 & BTS-DIDO & \first \textbf{0.029} & \first \textbf{0.680} & \first \textbf{0.074} & \first \textbf{4.446} & \first \textbf{99.8} & \first \textbf{99.7} & 0.041 & \multicolumn{2}{c}{} \\ 
\cmidrule{1-9}
\multirow{4}{*}{5} & Org + AuxUE & 0.059 & 1.760 & 0.067 & 4.496 & \second {98.5} & \second {97.1} & \second {0.035} & \multicolumn{2}{c}{} \\
 & BTS-SinglePU & \second {0.045} & 1.202 & 0.075 & 4.831 & 87.5 & 64.6 & \first \textbf{0.005} & \multicolumn{2}{c}{} \\
 & BTS-DEns. & \first \textbf{0.036} & \first \textbf{0.890} & \second {0.080} & \first \textbf{5.044} & 92.2 & 70.4 & \first \textbf{0.005} & \multicolumn{2}{c}{} \\
 & BTS-DIDO & \first \textbf{0.036} & \second {1.010} & \first \textbf{0.084} & \second {4.964} & \first \textbf{99.5} & \first \textbf{99.3} & 0.051 & \multicolumn{2}{c}{} \\
\cmidrule[\heavyrulewidth]{1-9}
\end{tabular}
}
\caption{\textbf{Ablation study on the necessity of using AuxUE.} Epistemic uncertainty estimation performance comparison on KITTI and KITTI-C. On clean KITTI, the extra columns stand for the dataset change experiment.}
\label{tab:abla_btsdido_2}
\end{table*}

\begin{figure*}[t]
        \begin{subfigure}{\textwidth}
        \centering
        A\includegraphics[width=0.7\linewidth]{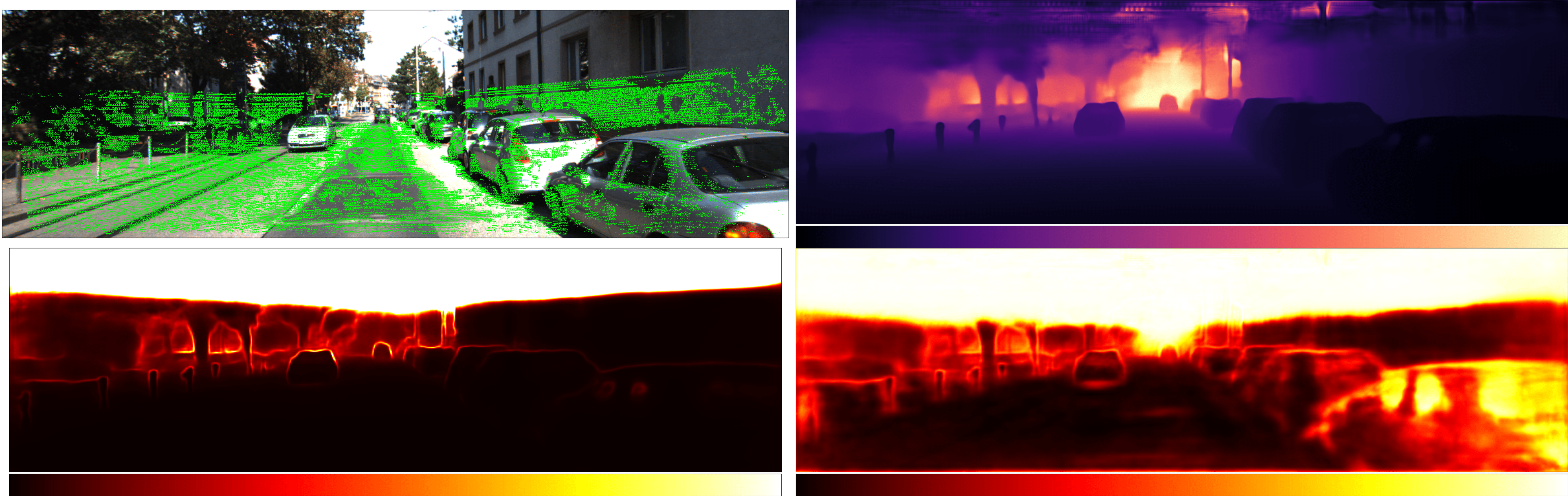}
\vspace{1em}
        \end{subfigure}
        \begin{subfigure}{\textwidth}
        \centering
        B\includegraphics[width=0.7\linewidth]{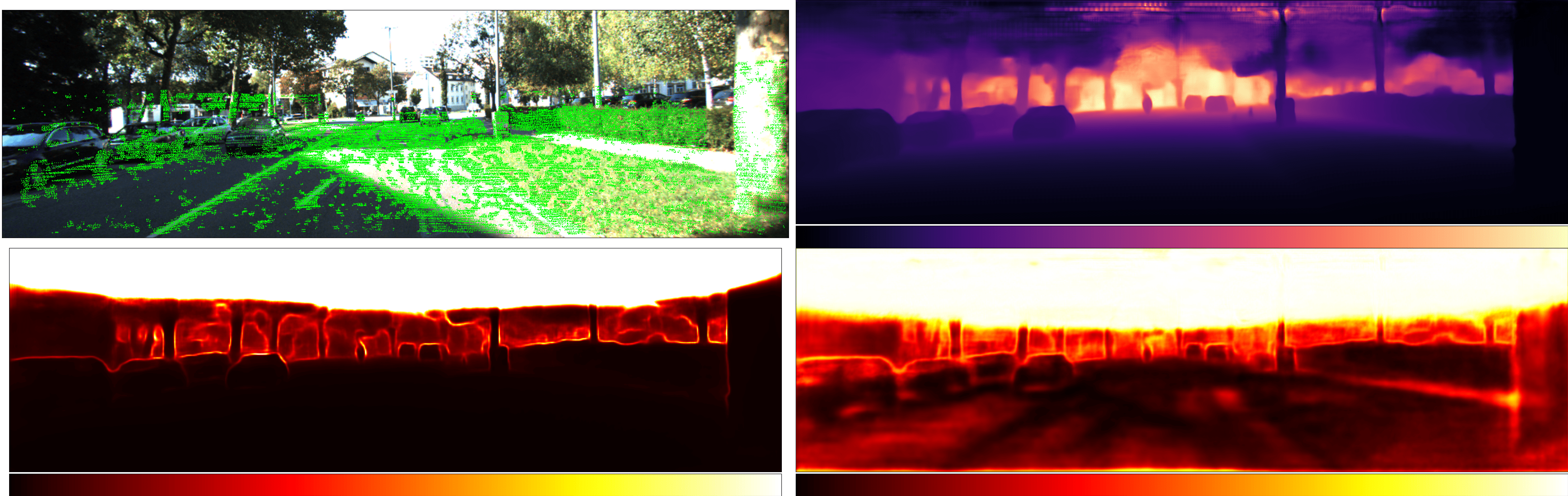}
\vspace{1em}
        \end{subfigure}
        \begin{subfigure}{\textwidth}
        \centering
        C\includegraphics[width=0.7\linewidth]{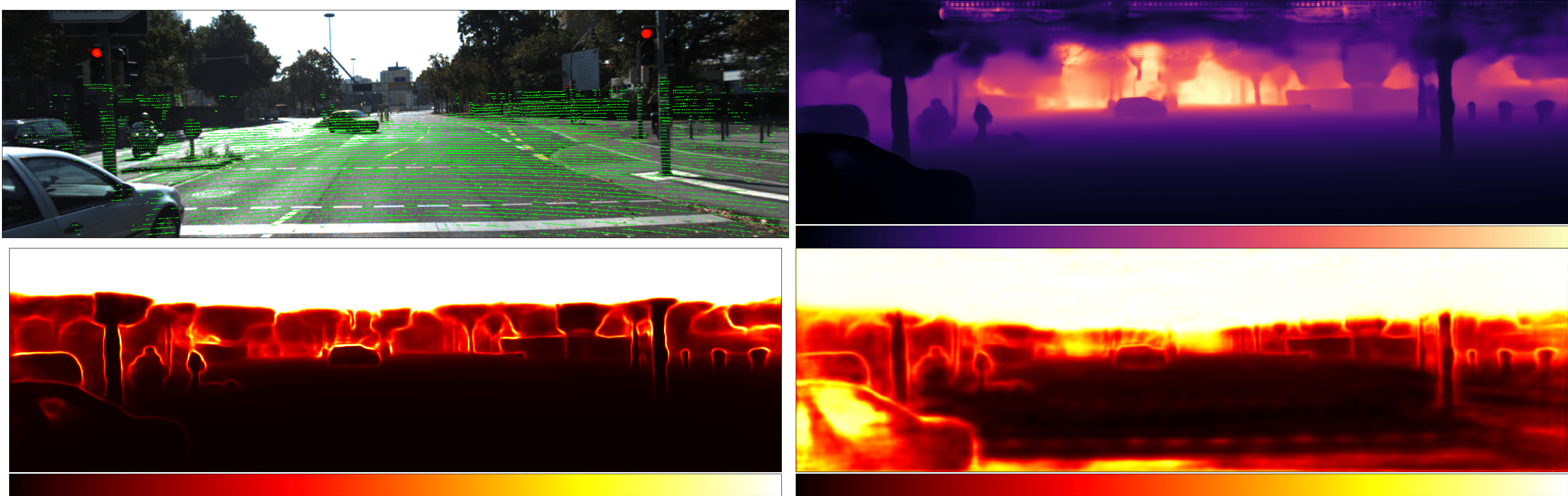}
        \end{subfigure}
\end{figure*}

\begin{figure*}[t]\ContinuedFloat
    \begin{subfigure}{\textwidth}
        \centering
        D\includegraphics[width=0.7\linewidth]{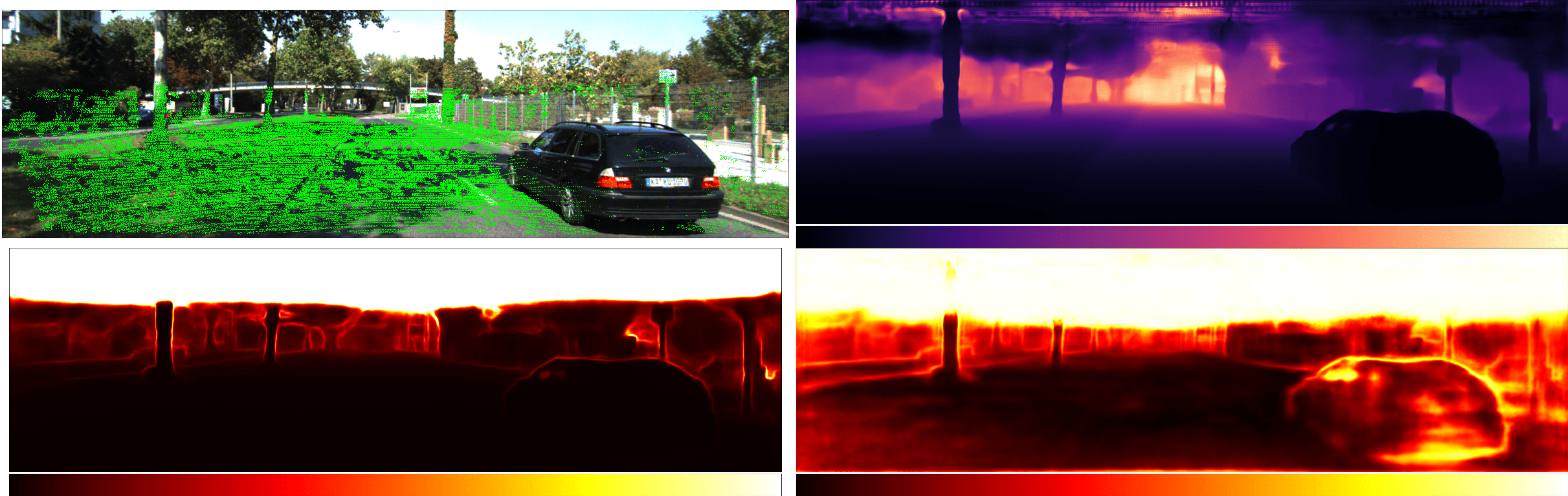}
\vspace{1em}
    \end{subfigure}
    \begin{subfigure}{\textwidth}
        \centering
        E\includegraphics[width=0.7\linewidth]{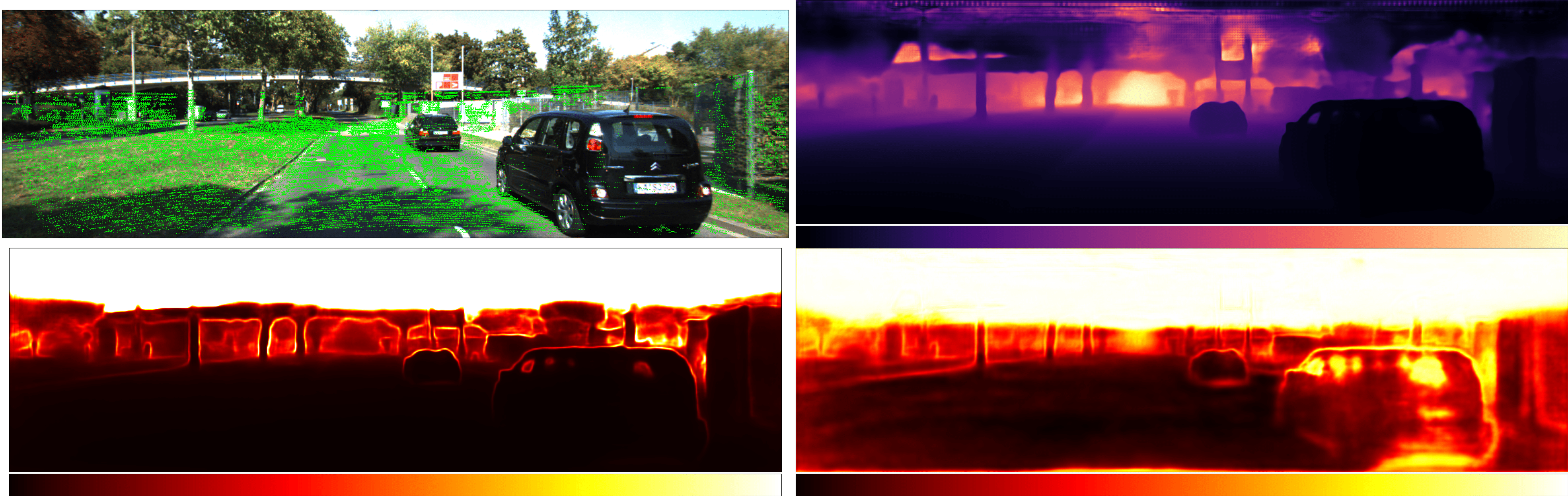}
\vspace{1em}
    \end{subfigure}
    \begin{subfigure}{\textwidth}
        \centering
        F\includegraphics[width=0.7\linewidth]{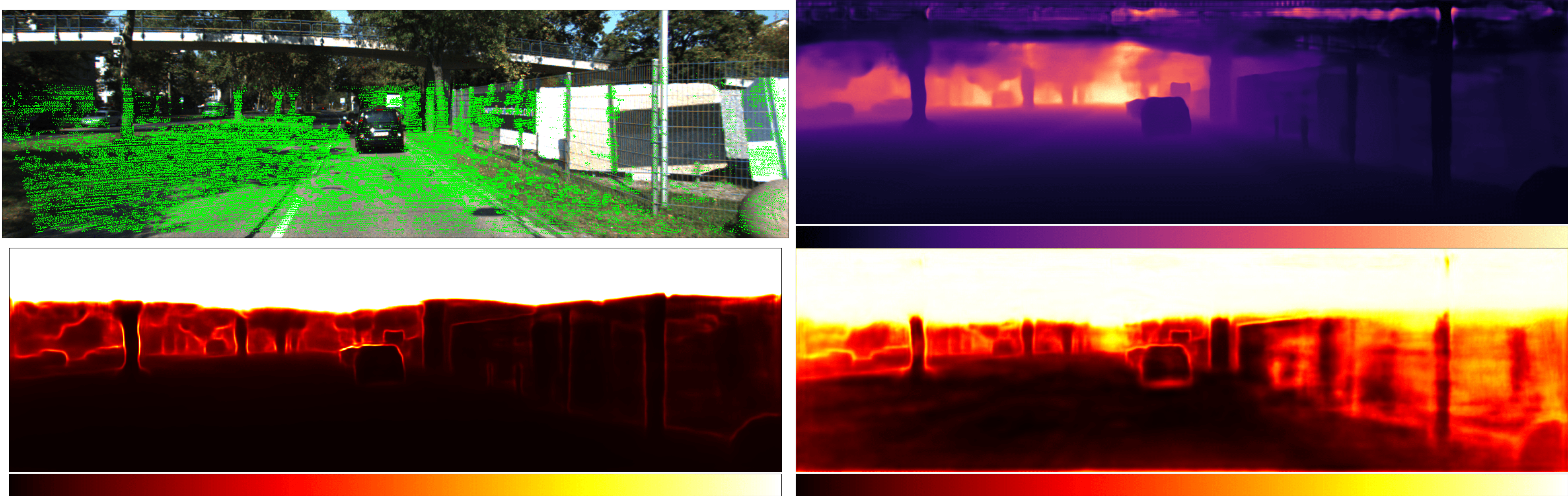}
    \end{subfigure}
\caption{\textbf{Visualizations on monocular depth estimations and corresponding uncertainty quantification results.} The color bars and the image orders follow the ones in  Fig.3 of the main paper.}
\label{tab:vis_mde_supp}
\end{figure*}

\clearpage